\documentclass[sigconf]{acmart}
\usepackage{amsmath}

\usepackage{amssymb}
\usepackage{gensymb}
\usepackage{booktabs}
\usepackage[inline]{enumitem}
\usepackage{libertine}
\usepackage{pifont}

\newcommand{\ncmark}{\ding{51}}
\newcommand{\nxmark}{\ding{55}}
\AtBeginDocument{%
  \providecommand\BibTeX{{%
    \normalfont B\kern-0.5em{\scshape i\kern-0.25em b}\kern-0.8em\TeX}}}

\copyrightyear{2020} 
\acmYear{2020} 
\setcopyright{acmcopyright}\acmConference[MM '20]{Proceedings of the 28th ACM International Conference on Multimedia}{October 12--16, 2020}{Seattle, WA, USA}
\acmBooktitle{Proceedings of the 28th ACM International Conference on Multimedia (MM '20), October 12--16, 2020, Seattle, WA, USA}
\acmPrice{15.00}
\acmDOI{10.1145/3394171.3413868}
\acmISBN{978-1-4503-7988-5/20/10}

\acmSubmissionID{911}

\begin{document}
\fancyhead{}

\title{Controllable Continuous Gaze Redirection}

\author{
Weihao Xia$^{1}$, 
Yujiu Yang$^{1}$, 
Jing-Hao Xue$^{2}$, %
Wensen Feng$^{3}$
}

\affiliation{
$^1$Tsinghua Shenzhen International Graduate School, Tsinghua University \\
$^2$Department of Statistical Science, University College London \\
$^3$Beijing University of Chemical Technology}

\email{xiawh3@outlook.com,  yang.yujiu@sz.tsinghua.edu.cn,  jinghao.xue@ucl.ac.uk,  sanmumuren@126.com}

\begin{abstract}
In this work, we present \textit{interpGaze}, a novel framework for controllable gaze redirection that achieves both precise redirection and continuous interpolation. 
Given two gaze images with different attributes, our goal is to redirect the eye gaze of one person into any gaze direction depicted in the reference image or to generate continuous intermediate results. To accomplish this, we design a model including three cooperative components: an encoder, a controller and a decoder. The encoder maps images into a well-disentangled and hierarchically-organized latent space. The controller adjusts the magnitudes of latent vectors to the desired strength of corresponding attributes by altering a control vector. The decoder converts the desired representations from the attribute space to the image space.
To facilitate covering the full space of gaze directions, we introduce a high-quality gaze image dataset with a large range of directions, which also benefits researchers in related areas.
Extensive experimental validation and comparisons to several baseline methods show that the proposed \textit{interpGaze} outperforms state-of-the-art methods in terms of image quality and redirection precision. 
\end{abstract}



\begin{teaserfigure}
\centering
  \includegraphics[width=0.85\textwidth]{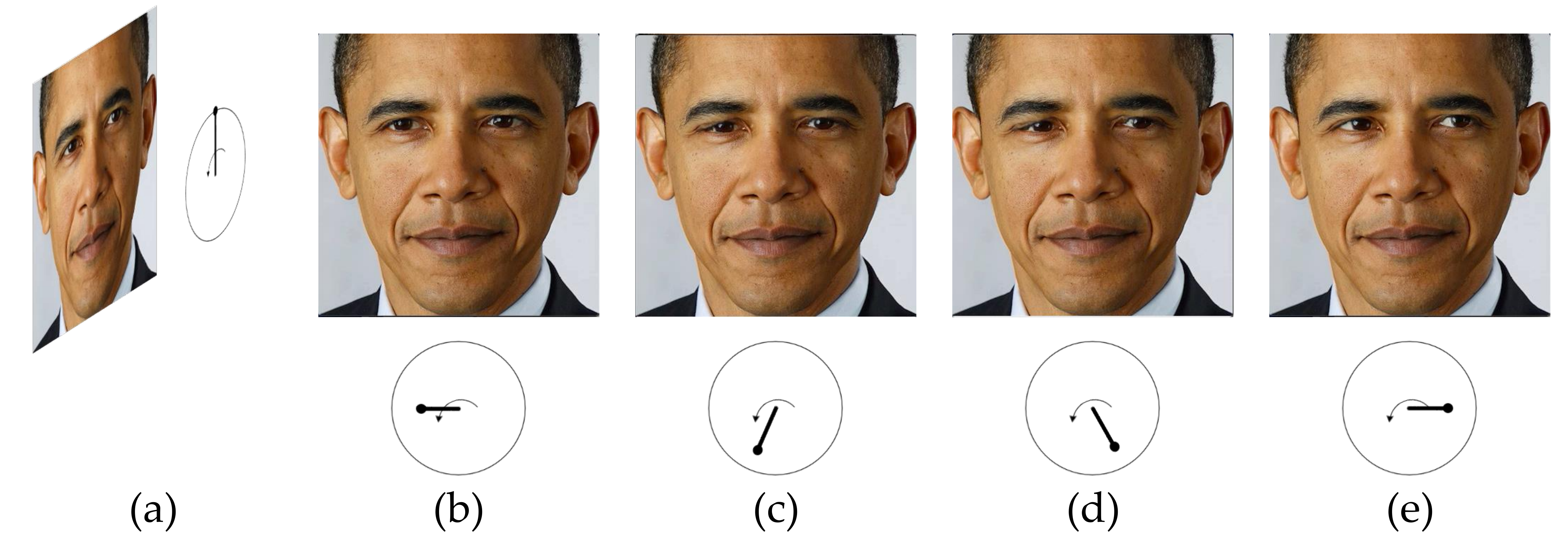}
  \caption{Controllable Continuous Gaze Redirection. (a) A sample image of Barack Obama and the gaze direction space definition; (b-e) Some gaze redirection results from the proposed \textit{interpGaze}. The dots are target directions, and the circles indicate the possible values in the full space of gaze directions.}
  \label{fig:teaser}
\end{teaserfigure}

\maketitle

\section{Introduction}
\label{sec:intro}

Eye contact plays a crucial role in our daily social communication since it conveys important non-verbal cues such as emotion, intention and attention.
Gaze redirection is an emerging research topic in computer vision and computer graphics, aiming at manipulating the gaze of a given image to the desired direction with an angle or image as the reference, as illustrated in Figure~\ref{fig:teaser}.
This task is important in many real-world scenarios. For example, the case that people are not looking at the camera at the same time occurs frequently when taking a group photo. In the video conference, the participants do not have the chance to make direct eye contact due to the location disparity between the video screen and the camera. 
In both cases, gaze redirection can adjust each person’s eye gaze to the same direction or along a single direction to simulate eye contact.
Gaze redirection technique could also alleviate the data insufficiency problem for gaze estimation in the wild, by synthesizing novel samples to augment existing datasets.

Traditional methods typically re-render the entire input region by performing 3D transformations, which requires heavy instrumentation to acquire the depth information~\cite{kuster2012gaze, yang2002eye, zhu2011eye, criminisi2003gaze}.
Learning-based gaze redirection approaches~\cite{ganin2016deepwarp, he2019gaze} has risen in recent years. For example, DeepWarp~\cite{ganin2016deepwarp} uses a neural network to warp the input image to the desired direction by predicting the dense flow field. Their method fails to generate photo-realistic images for large redirection angles, especially in the presence of large dis-occlusions, such as large parts of the eyeball being covered by the eyelid in the source image.
More importantly, such warping methods cannot generate precise gaze redirection, as it only minimizes the pixel-wise differences between the synthesized and ground-truth images without any geometric regularization.

\begin{figure}[t]
\begin{center}
\includegraphics[width=.85\linewidth]{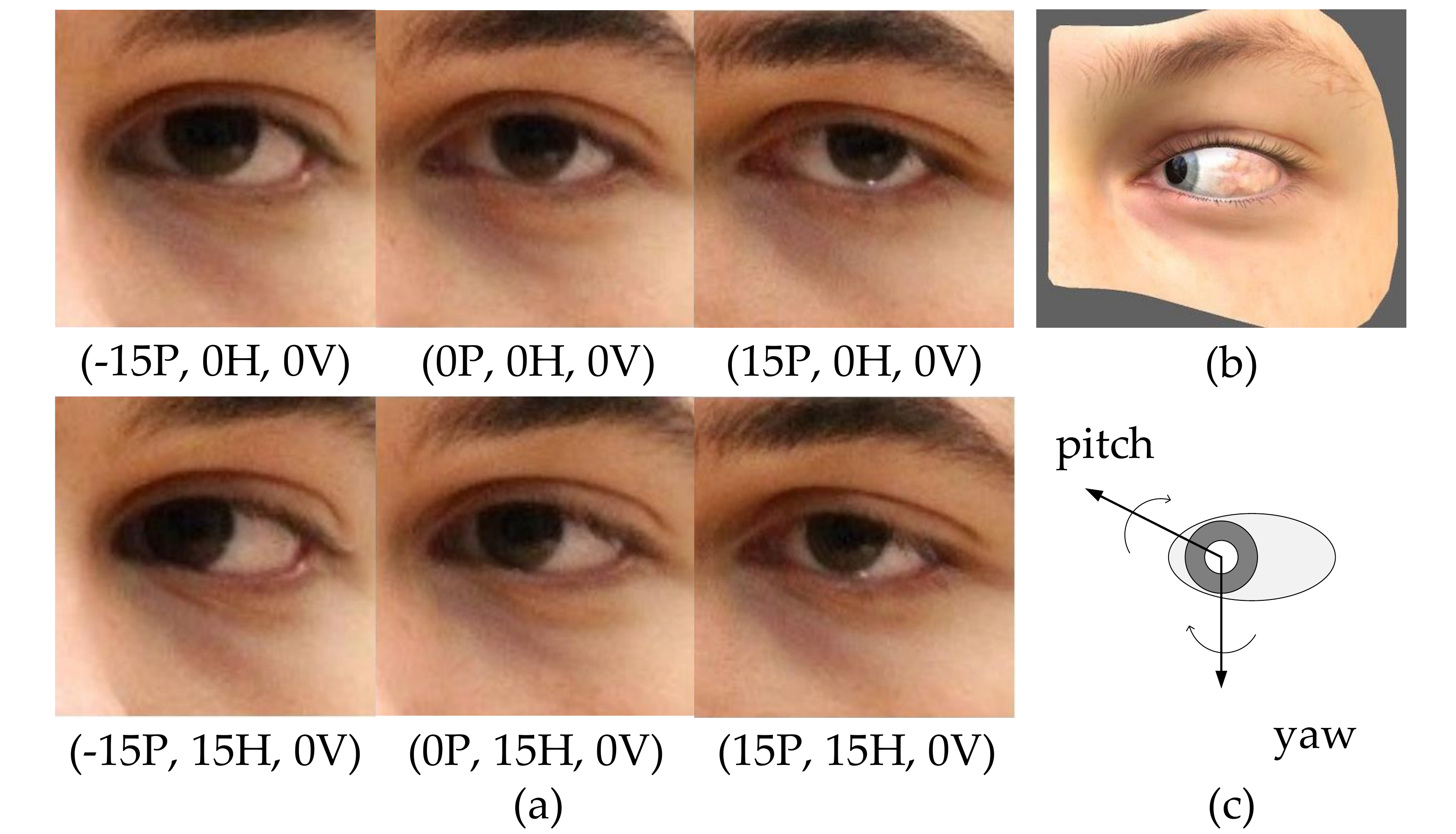}
\end{center}
\caption{\textbf{Attribute illustration}. (a) Samples from the Columbia Gaze~\cite{smith2013gaze}. Typically, gaze-related attributes are head pose ($\textbf{P}$), horizontal gaze direction (yaw, $\textbf{H}$) and vertical gaze direction (pitch, $\textbf{V}$). The directions are illustrated in (c). Number means angle, and negative sign means left hand direction of the subject. Each row of (a) demonstrates the same gaze direction at different head poses, and each column shows the gaze difference at the same head pose. The positions of eye balls with the same gaze direction are distinct at different head poses. (b) A sample from the synthetic UnityEyes~\cite{wood2016UnityEyes}. (c) Gaze direction illustration.}
\label{fig:attribute_illustration}
\end{figure}

He \textit{et al.} \cite{he2019gaze} first introduce Generative Adversarial Network (GAN) to synthesize photo-realistic eye images conditioned on a target gaze direction.
Since they incorporate both source gaze image and target gaze vector as input and use a high-quality gaze dataset~\cite{smith2013gaze} for training, their method can synthesize images with high quality and redirection precision.
However, they do not incorporate head pose together with the vertical and horizontal gaze direction. Head pose is also an important attribute for gaze redirection since the positions of eye balls with the same gaze direction are distinct at different head poses as shown in Figure~\ref{fig:attribute_illustration}(a). That means they need to train a model for each head pose to precisely control the redirection. 
Furthermore, although able to produce photo-realistic and precise redirection images, the vector-based methods are prone to the gaze ranges defined in the training dataset and fail to generalize to unseen angles, 
which hinders the models to cover the full space of gaze directions. 
Odobez \textit{et al.}~\cite{odobez2019gaze} propose a self-supervised method to adapt the model pretrained from large amounts of well-aligned synthetic data to real data. Their method produces continuous but perceptually implausible results, due to the limitation of synthetic data as shown in Figure~\ref{fig:attribute_illustration}(b).

\begin{figure*}[ht]
\begin{center}
\includegraphics[width=0.95\linewidth]{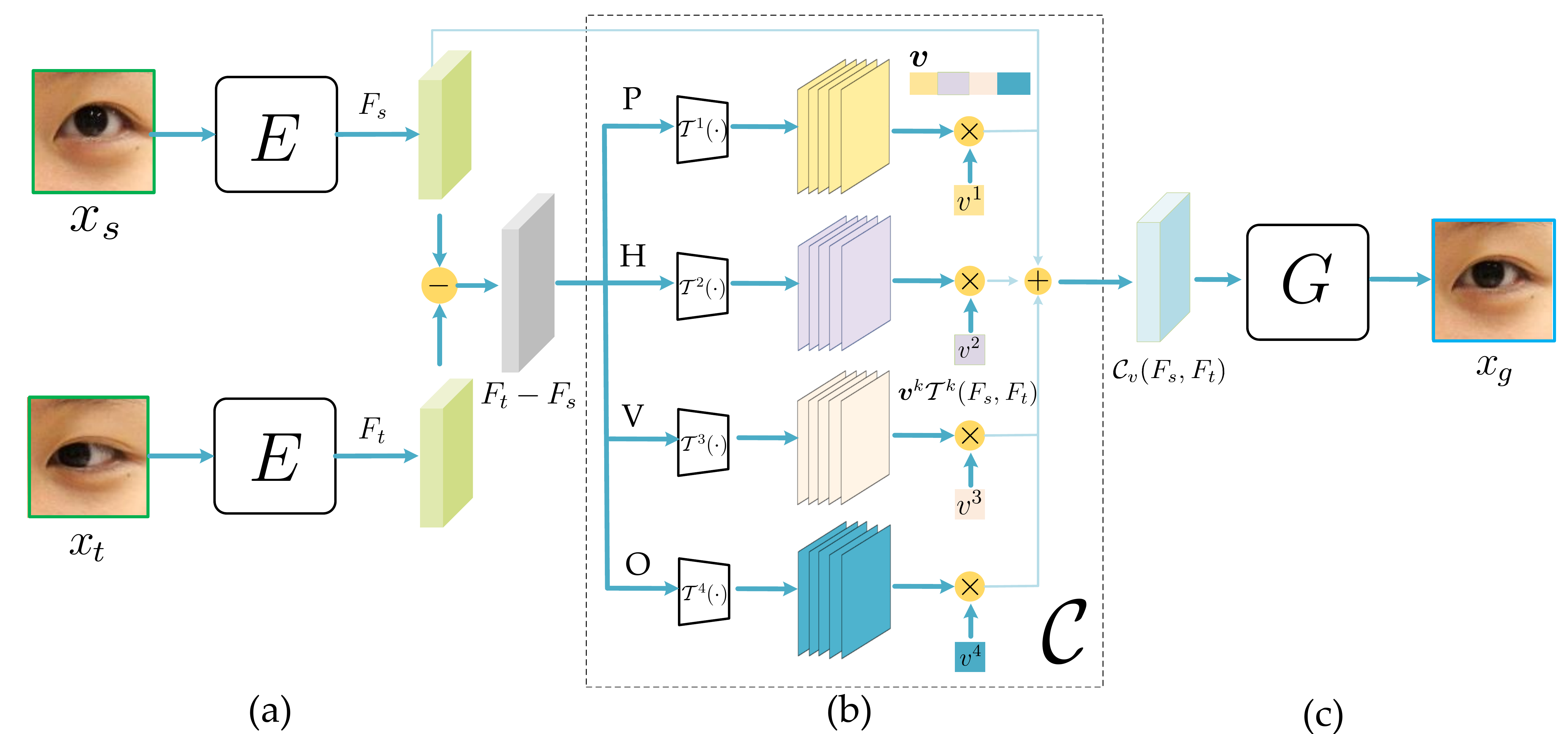}
\end{center}
\caption{\textbf{Overview of the proposed \textit{interpGaze}.} Our proposed model contains (a) an \textbf{Encoder} $\boldsymbol{{E}}$, (b) a \textbf{Controller} $\boldsymbol{\mathcal{C}}$ and (c) a \textbf{Decoder} $\boldsymbol{G}$. The Encoder $\boldsymbol{E}$ maps images $\boldsymbol{x}_s$ and $\boldsymbol{x}_t$
into feature space $F_{s}=\boldsymbol{E}\left(x_{s}\right)$ and  $F_{t}=\boldsymbol{E}\left(x_{t}\right)$. Then the feature difference is fed into four branches of the controller $\boldsymbol{\mathcal C}$ to produce morphing results of two samples $\mathcal {C}_{\boldsymbol{v}}(F_{s},F_{t}) =F_{s}+\sum_{k=1}^{c+1}{\boldsymbol{v}}^{k} \mathcal{T}^{k}(F_{t}-F_{s})$. 
The abbreviations P, H, V and O are head pose (P), vertical gaze direction (pitch, V), horizontal gaze direction (yaw, H) and miscellaneous attributes.
The ``O'' branch is designed for other secondary attributes like glass, eyebrow, skin color, hair and illumination. The control vector $\boldsymbol{v} \in[0,1]^{(c+1) \times 1}$ adjusts the strength of each attribute, where $c=3$ in current setting. The Decoder $\boldsymbol{G}$ maps the latent features back to the image space. Please refer to Section~\ref{sec:method} for details.}
\label{fig:overview}
\end{figure*}

To address the limitations of previous methods, \textit{i.e.}, (1) low-quality generation, (2) low redirection precision and (3) gaze angle limitation, we propose a novel controllable gaze redirection method that can achieve precise redirection and continuous interpolation in one model.
More specifically, our method works on both eye image synthesis conditioned on target gaze directions, and continuous intermediate gaze change between two given gaze images. 
As shown in Figure~\ref{fig:overview}, our model contains an Encoder $\boldsymbol{E}$, a Controller $\boldsymbol{\mathcal{C}}$ and a Decoder $\boldsymbol{G}$. The Encoder $\boldsymbol{E}$ maps two input gaze images, source image $\boldsymbol{x}_s$ and target image $\boldsymbol{x}_t$, into a latent feature space with $F_s=\boldsymbol{E}(x_s)$ and $F_t=\boldsymbol{E}(x_t)$. 
Then the feature difference between $F_s$ and $F_t$ is fed into four branches of the controller $\boldsymbol{\mathcal {C}}$ to produce morphing results of two samples $\mathcal {C}_{\boldsymbol{v}}(F_s,F_t)$. 
The four branches represent head pose, gaze yaw, pitch and miscellaneous attributes respectively. The Decoder $\boldsymbol{G}$ maps the latent features back to the image space. 
To achieve precise redirection and continuous interpolation, we introduce a gaze vector $\boldsymbol{v} \in [0,1]^{c \times 1}$ to the controller $\boldsymbol{\mathcal {C}}$. Each element of $\boldsymbol{v}$ corresponds to a mixing indicator for each attribute.
Since the head pose, gaze yaw, gaze pitch and miscellaneous attributes have been well-disentangled and hierarchically-organized in the latent space, we can adjust each attribute by altering the control vector $\boldsymbol{v}$.
For precise gaze redirection, the control vector $\boldsymbol{v}$ selects the features of target directions to generate photo-realistic eye images with the desired head pose and gaze directions.    
For continuous interpolation, the control vector $\boldsymbol{v}$ interpolates gradually in the latent space of the difference between the source gaze images and the target gaze images. More importantly, benefiting from the flat and smooth latent space, we can generate interpolation sequences in different orders by assigning $\boldsymbol{v}$, and can generalize to exaggeration of certain attribute by setting corresponding dimension $v^k$ to be a reasonable value greater than 1.
The two completely different tasks are unified into the same framework through the proposed feature disentanglement and control mechanism, while avoiding the deficiencies in covering the gaze space of vector-based redirection methods.

In addition, to alleviate the current drawbacks of existing datasets, we collect a large-scale high-quality dataset for gaze redirection and other related tasks, which can also benefit research in related areas. It covers a larger range of eye angles than other popular gaze redirection dataset like Columbia Gaze~\cite{smith2013gaze} and is also with diversity on eye shapes, glasses, ages and genders. 

Our main contributions can be summarized as follows:
\begin{itemize}
    \item We present \textit{interpGaze}, a novel framework achieving both precise gaze redirection and continuous gaze interpolation. The two different tasks can be readily controlled by altering a control vector.%
    \item We learn a well-disentangled and hierarchically-organized latent space by decoupling the related gaze attributes and equipping with the efficacy of one-shot and diversity, which makes our method interpretable and complete.
    \item We contribute a high-quality gaze dataset, which contains a large range of gaze directions and diversity on eye shapes, glasses, ages and genders, to benefit other researchers in related areas.
\end{itemize}

\section{Related Work}
\label{related_work}

\vspace*{2mm}\noindent {\bf Gaze Redirection.} 
Some traditional approaches are based on 3D modeling~\cite{banf2009example, wood2018gazedirector}. They use 3D morphable models to fit both texture and shape of the eye, and re-render the synthesized eyeballs superimposed on the source image. However, these methods make strong model assumptions that may not hold in practice. Therefore, they can not handle images with eyeglasses and other high-variability inter-personal differences. Some others~\cite{kuster2012gaze, yang2002eye, zhu2011eye, criminisi2003gaze} render a scene containing the face of a subject from a given viewpoint to mimic gazing at the camera. These methods require a depth map and synthesize a person-specific image with the redirected gaze by performing 3D transformations.

Learning-based methods have shown remarkable results from using a large dataset labelled with head pose and gaze angles.
He \textit{et al.}~\cite{he2019gaze} first introduce Generative Adversarial Network (GAN) to generate photo-realistic eye images while preserving the desired gaze direction. However, the lack of large redirection angles hinders the models to cover the full space of gaze directions.
Odobez \textit{et al.}~\cite{odobez2019gaze} propose a self-supervised method to adapt the model pretrained from large amounts of well-aligned synthetic data to real data.
Their method produces continuous but perceptually implausible results.
There are also some face manipulation methods~\cite{choi2018stargan,Iizuka2017completion} based on image translation or image inpainting to redirect gaze but lack of the capability to precisely control the transformation.

\vspace*{2mm}\noindent {\bf Continuous Image Generation.}
Recently, there has been some studies on continuous image generation. Most of them can be divided into two categories: 
\begin{enumerate*}[label=(\alph*)] 
\item Learning explicit or implicit 3D representations and rendering images in different viewpoints. 
For example, Wood \textit{et al.}~\cite{wood2018gaze} fit a parametric eye region model to images and perform gaze redirection by warping eyelids, and compositing eyeballs onto the output in a photo-realistic manner.
Chen \textit{et al.}~\cite{chen2019mono} present a transforming auto-encoder in combination with a depth-guided warping procedure that learns to produce geometrically accurate views with fine-grained (\textit{e.g.}, $1\degree$ step-size) camera control for both single objects and natural scenes.
Olszewski \textit{et al.}~\cite{olszewski2019tbn} propose a \textit{Transformable Bottleneck Networks} to learn the 3D spatial structure from a single image for arbitrary novel-view synthesis. Mildenhall \textit{et al.}~\cite{mildenhall2020scenes} represent scenes as \textit{Neural Radiance Fields} for synthesizing novel views from a sparse set of input views.

\item Learning continuous 2D latent features. For example, Pumarola \textit{et al.}~\cite{Pumarola_ijcv2019} introduce a novel GAN model conditioned on a continuous embedding of muscle movements defined by Action Units (AU) annotations, which is able to generate novel facial expressions in a continuum. But their method requires continuous label annotation.
Chen \textit{et al.}~\cite{chen2019Homomorphic} propose an unsupervised image-to-image translation framework aiming at generating naturally and gradually changing intermediate facial images.
Lira \textit{et al.}~\cite{lira2020hop} design a multi-hop mechanism that transforms images gradually between two input domains without any in-between hybrid training images. 
More recently, Abdal \textit{et al.}~\cite{abdal2019embed} propose to directly embed two given images into the latent space of StyleGAN~\cite{kerras2019stylegan}, and compute the morphing $\omega$ based on a linear interpolation of the obtained vectors $\omega_1$, $\omega_2$ by $\omega = \lambda \cdot \omega_1+(1-\lambda) \cdot \omega_2$, where $\lambda\in(0,1)$ controls the level of mixing of two samples.
\end{enumerate*}

\section{Methodology}
\label{sec:method}
\subsection{Overview}
Our goal is to learn a model which can achieve both {\bf precise redirection} and {\bf continuous interpolation}. 
For an RGB image of an eye patch $x \in \mathbb{R}^{H \times W \times 3}$, we define three primary attributes, \textit{i.e.}, gaze direction $\boldsymbol{d}_g$ and head pose $\boldsymbol{d}_h$, where $\boldsymbol{d}_g = [\phi_g, \theta_g]$, $\phi_g \in \mathbb{R}$ and $\theta_g \in \mathbb{R}$ denote the target yaw and pitch angles, respectively. Given two gaze images $x_s$ and $x_t$ with different attributes, the task is to redirect the eye gaze of one person $x_s$ into any gaze direction $x_g$ depicted in the reference image of another person $x_t$, or to generate continuous intermediate results $x_g$ between $x_s$ and $x_t$ of the same person.  

To achieve this, we design a model including an encoder $\boldsymbol{E}$, a controller $\boldsymbol{\mathcal{C}}$ and a decoder $\boldsymbol{G}$. The encoder $\boldsymbol{E}$ maps images ${x}_s$ and ${x}_t$ into feature space $F_{s}=\boldsymbol{E}(x_{s})$ and  $F_{t}=\boldsymbol{E}(x_{t})$. The controller $\boldsymbol{\mathcal C}$ produces morphing results of two samples $\mathcal C(F_{s},F_{t})$. The decoder $\boldsymbol{G}$ maps the desired features back to the image space.
Figure \ref{fig:overview} provides a full overview of the method. The precise redirection and continuous interpolation can be achieved using the same model by altering a control vector $\mathcal{\boldsymbol{v}}$. We elaborate on the three components in more detail below.

\begin{figure}[t]
\begin{center}
\includegraphics[width=1.0\linewidth]{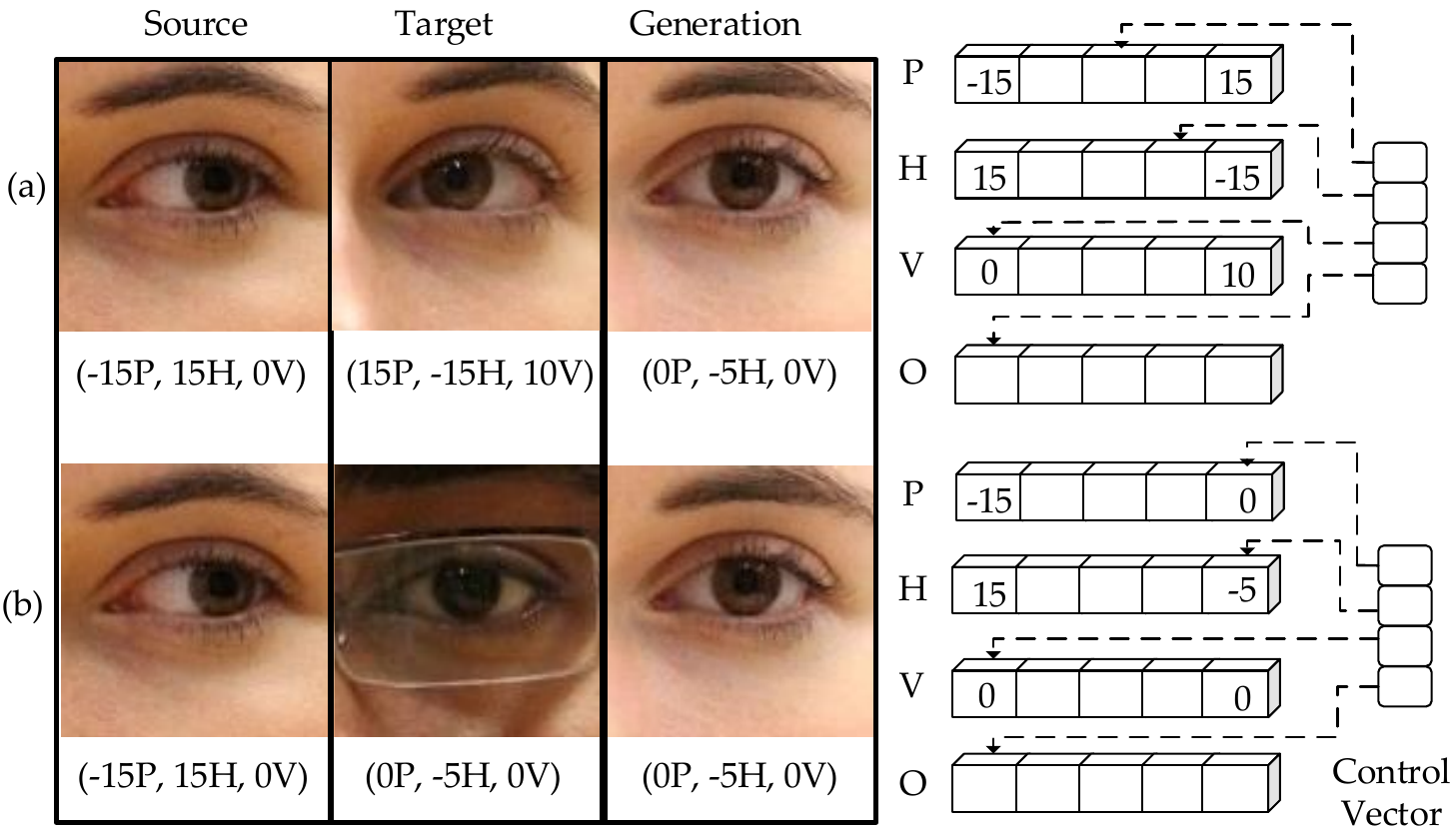}
\end{center}
\caption{\textbf{Illustration of Control Mechanism of \textit{interpGaze} for (a) gaze interpolation and (b) gaze redirection.} 
The two completely different tasks are unified into the same framework and can be controlled by altering a control vector.
Please refer to Section~\ref{sec:control_mechanism} for more details.
}
\label{fig:control_mechanism}
\end{figure}

\subsection{Encoder and Decoder}
To make the precise redirection and continuous interpolation feasible, we need the encoder $\boldsymbol{E}$ unfold the natural image manifold to a flat and smooth structure of latent space. We use WGAN-GP \cite{gulrajani2017improved} to train our model due to its stable performance. 
The encoder $\boldsymbol{E}$ and the controller $\boldsymbol{\mathcal{C}}$ are trained to minimize the Wasserstein distance between real gaze images and generated ones, while a discriminator $\boldsymbol{\mathcal{D}}$ is trained to maximize the distance. It can be formulated as
\begin{equation}
\label{eqn:adv_loss}
\begin{aligned}
&\min _{\boldsymbol{\mathcal{D}}} \mathcal{L}_{GAN_{\boldsymbol{\mathcal{D}}}}=\mathbb{E}_{\mathbb{P}_{f}}[\boldsymbol{\mathcal{D}}(\hat{F})]-\mathbb{E}_{\mathbb{P}_{r}}[\boldsymbol{\mathcal{D}}(F)]+\lambda_{gp} \mathcal{L}_{gp},\\
&\min_{\boldsymbol{E}, \boldsymbol{\mathcal{C}}} \mathcal{L}_{GAN_{\boldsymbol{E}, \boldsymbol{\mathcal{C}}}}=\mathbb{E}_{\mathbb{P}_{r}}[\boldsymbol{\mathcal{D}}(F)]-\mathbb{E}_{\mathbb{P}_{f}}[\boldsymbol{\mathcal{D}}(\hat{F})].
\end{aligned}
\end{equation}
In Equation~\ref{eqn:adv_loss}, $F={E}(x)$ is the feature extracted from gaze image $x$ by the encoder $\boldsymbol{E}$; $\hat{F}=\mathcal{C}(F_{i}, F_{j})$ is the mixing feature generated by a morphing function of the obtained vectors $F_{i}, F_{j}$; 
$\mathbb{P}_{r}$ and $\mathbb{P}_{f}$ are the distributions of real and fake gaze images respectively; and $\mathcal{L}_{gp}$ is the gradient penalty term~\cite{gulrajani2017improved} used to maintain the Lipschitz continuity of $\boldsymbol{\mathcal{D}}$. 
The hyperparameter $\lambda_{gp}$ controls the strength of the gradient penalty, and we use $\lambda_{gp}=10$ in all experiments. 
Here the encoder $\boldsymbol{E}$ works together with the controller $\boldsymbol{\mathcal{C}}$ to generate target gaze images. 
We will introduce more details of controller $\boldsymbol{\mathcal{C}}$ in the next section. \par

We additionally incorporate a decoder $\boldsymbol{G}$ to invert features back to images. The decoder $\boldsymbol{G}$ is trained with perceptual loss and  reconstruction loss.

\begin{figure*}[ht]
\begin{center}
\includegraphics[width=.85\linewidth]{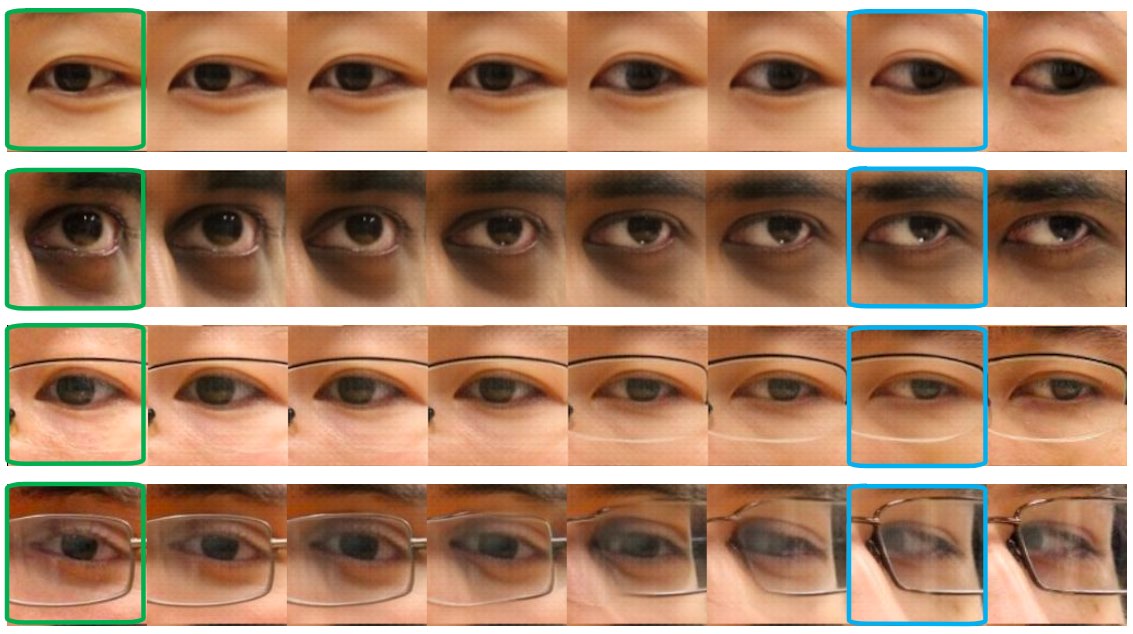}
\end{center}
\caption{\textbf{Illustration of gaze interpolation and extrapolation. The four examples are produced by the proposed \textit{interpGaze}.} {\textcolor{green}{Green}} rectangles are the start gaze images and {\textcolor{blue}{blue}} ones are the ends. Between them are intermediate results of the two given gaze images. The last column is the result of extrapolation. Detailed interpretation can be found in Section~\ref{sec:interp_results}.}
\label{fig:interpolation}
\end{figure*}

The perceptual loss proposed in~\cite{johnson2016perceptual} penalizes the decoder $\boldsymbol{G}$ for generating images that match ground-truth images perceptually. Typically, the perceptual loss is defined by the VGG-16~\cite{simonyan2014very} model pre-trained on ImageNet \cite{krizhevsky2012imagenet}. 
The perceptual loss contains two terms, the content loss $\mathcal{L}_{c}$ and style loss $\mathcal{L}_{s}$, which are defined as
\begin{equation}
\begin{aligned}
\min _{\boldsymbol{G}} \mathcal{L}_{c} &=\mathbb{E}\left(\frac{1}{H_{j} W_{j} C_{j}}\left\|\psi_{j}(\boldsymbol{G}(F))-\psi_{j}(x_s)\right\|^{2}\right),
\end{aligned}
\end{equation}

\begin{equation}
\begin{aligned}
\min_{\boldsymbol{G}}\mathcal{L}_{s} &=\boldsymbol{\mathbb{E}}\left(\sum_{j=1}^{J}\left\|f_{j}(\boldsymbol{G}(F))-f_{j}(x_s)\right\|^{2}\right),
\end{aligned}
\end{equation}
where $\psi$ denote the pre-trained VGG-$16$ network, $\psi_{j}(\boldsymbol{x}) \in \mathbb{R}^{H_{j} \times W_{j} \times C_{j}}$ is the activation of $j$-th layer of $\psi$, $\mathcal{L}_{s}$ is the sum of all style losses from the 1-st layer to the $J$-th layer of the VGG model. $f_{j}(\boldsymbol{x})$ denotes the Gram matrix, the entry of which is defined as
\begin{equation}
\begin{aligned}
f_{j}(\boldsymbol{x})_{c, c^{\prime}}=\frac{1}{H_{j} W_{j} C_{j}} \sum_{h}^{H_{j}} \sum_{w}^{W_{j}} \psi_{j}(\boldsymbol{x})_{h, w, c} \psi_{j}(\boldsymbol{x})_{h, w, c^{\prime}}.
\end{aligned}
\end{equation}
The perceptual loss is the sum of content loss and style loss: $\mathcal{L}_{p} = \mathcal{L}_{c} + \mathcal{L}_s$. The above two loss terms can force the generated eye patch images to be photo-realistic, and ensure redirection of the gaze directions simultaneously. 
Following \cite{zhu2017unpaired}, we enforce cycle consistency as the reconstruction term to ensure that personalized features are maintained during the redirection process, which is defined as
\begin{equation}
\begin{aligned}
\min_{\boldsymbol{E},\boldsymbol{G}} \mathcal{L}_{recon} &=\mathbb{E}(\left\|(\boldsymbol{G}(\boldsymbol{E}(x_g)))-x_s\right\|^{2}).
\end{aligned}
\end{equation}

We also introduce a knowledge distillation loss to guide the training of $\boldsymbol{E}$. 
Knowledge distillation~\cite{bengio2013distill} is widely used to compress CNN models and has recently been utilized for latent space interpolation~\cite{Ulyanov18GEN,chen2019Homomorphic}.
However, distilling the dark knowledge from the teacher’s output pixels is difficult. Instead, we match the intermediate representations as explored in prior studies~\cite{li2020gan}.
The intermediate layers allow the student model to acquire more information in addition to outputs, as they contain more channels and provide richer information.
The distillation objective can be formalized as
\begin{equation}
\begin{aligned}
\min_{\boldsymbol{E}, \boldsymbol{F}} \mathcal{L}_{{distill}}=\sum_{t=1}^{T}\left\|\boldsymbol{E}_{t}(\mathbf{x})-\boldsymbol{F}_{t}\left(\psi_{t}(\mathbf{x})\right)\right\|_{2},
\end{aligned}
\end{equation}
where $\boldsymbol{E}_{t}(\mathbf{x})$ and $\psi_{t}(\mathbf{x})$ are the intermediate feature activations of the $t$-th chosen layer in the encoder $\boldsymbol{E}$ and VGG models, and $T$ denotes the number of layers. A $1 \times 1$ learnable convolution layer $\boldsymbol{F}_{t}$ maps the features from the student model to the same number of channels in the features of the teacher model. We jointly optimize $\boldsymbol{E}_{t}$ and $\boldsymbol{F}_{t}$ to minimize the distillation loss $\mathcal{L}_{distill}$. 

The final objective function of the encoder $\boldsymbol{E}$ and decoder $\boldsymbol{G}$ is
\begin{equation}
\begin{aligned}
\mathcal{L}_{\boldsymbol{E}}&= \lambda_{GAN_{\boldsymbol{E}}}\mathcal{L}_{GAN_{\boldsymbol{E}, \boldsymbol{\mathcal{C}}}} + \lambda_{{\boldsymbol{E}}} \mathcal{L}_{recon} + \lambda_{distill}\mathcal{L}_{distill}, \\
\mathcal{L}_{\boldsymbol{G}}&=\lambda_{p}\mathcal{L}_{p}+\lambda_{\boldsymbol{G}} \mathcal{L}_{recon},\\
\mathcal{L}_{\boldsymbol{\mathcal{D}}}&= \lambda_{GAN_{\boldsymbol{\mathcal{D}}}} \mathcal{L}_{GAN_{\boldsymbol{\mathcal{D}}}}.\\
\end{aligned}
\end{equation}
We set all scalars $\lambda_{i}$ as 1 in our experiments.
\subsection{Controller}
The controller $\boldsymbol{\mathcal{C}}$ produces morphing results $\boldsymbol{\mathcal{C}}(F_{s},F_{t})$ of two features. Then the decoder $\boldsymbol{G}$ maps the morphed latent features back to the image space.
This morphing process typically can be represented as the \textit{Latent Space Interpolation} ~\cite{bengio2013distill,Berthelot19RRG,Upchurch17DFI,Chen18FFP}. Specifically, 
this process can be done linearly as
\begin{equation}
\label{eqn:linear_interp}
\begin{aligned}
\mathcal{C}\left(F_{i}, F_{j}\right)=F_{i}+\alpha\left(F_{j}-F_{i}\right),
\end{aligned}
\end{equation}
where $F_{i}$ and $F_{j}$ are two features of real samples, and $\alpha \in [0,1]$ is a parameter that controls the mixing strength of two samples. The second term $\alpha(F_{j}-F_{i})$ can also be seen as the relative offset or a shifting vector that points from $F_{i}$ towards $F_{j}$.
These interpolation methods can connect images of different attributes and produces intermediate results.
Nevertheless, they fail to achieve precise control and explain how these attributes are mixed.

The controller $\boldsymbol{\mathcal{C}}$ plays two roles: \begin{enumerate*}[label=(\alph*)] \item disentangle the representations of these three primary gaze-related attributes and other attributes, and \item manipulate them precisely.\end{enumerate*}

To obtain disentangled representation for each attribute, we introduce a function $\mathcal{I}(\cdot)$ that maps latent feature $F_{i}$ to an attribute vector $z_i$, \textit{i.e.}, $\mathcal{I}(F_{i}) = z_i$. The translation from the latent space to the attribute space can be performed by the following structure-preserving map (and vice versa), which is analogous to~\textit{isomorphism} in algebra~\cite{Wong1992isomorphism}:
\begin{equation}
\label{eqn:attribute_rep_1}
\begin{aligned}
\mathcal{I}\left(\mathcal{C}\left(F_{i}, F_{j}\right)\right)=\mathcal{C}^{\prime}\left(\mathcal{I}\left(F_{i}\right), \mathcal{I}\left(F_{j}\right)\right),%
\end{aligned}
\end{equation}
where $\mathcal{C}^{\prime}\left(\boldsymbol{z}_{i}, \boldsymbol{z}_{j}\right)$ can be viewed as an morphing function that interpolates in the attribute space.

To manipulate these attributes flexibly,
we need a control vector $\boldsymbol{v} \in[0,1]^{c \times 1}$ to adjust the strength of each selected attribute.
With both extensions of Equation~\ref{eqn:linear_interp}, the revised version of morphing process $\mathcal{C}\left(F_{i}, F_{j}\right)$ can be written as a more flexible formulation 
$\mathcal{C}_{\boldsymbol{v}}\left(F_{i}, F_{j}\right)$.
The linear interpolation defined in Equation~\ref{eqn:linear_interp} is extended to a piece-wise one of
\begin{equation}
\label{eqn:homo_interp}
\begin{aligned}
\mathcal{C}_{\boldsymbol{v}}(F_{i}, F_{j})=F_{i}+\sum_{k=1}^{c+1} \boldsymbol{\boldsymbol{v}}^{k} \mathcal{T}^{k}(F_{j}-F_{i}),
\end{aligned}
\end{equation}
where $\mathcal{T}^{k}(\cdot)$ is a learnable mapping function represented by CNN, $\boldsymbol{v}^{k}$ is the $k$-th dimension of $\boldsymbol{v}$, $k=1, \cdots, c, c+1$, $c$ is the number of the three primary attributes (head pose, gaze yaw or pitch). We also add another branch for other secondary attributes such as glass, skin color, eyebrow and hair.

$\boldsymbol{v}^{k}$ corresponds to the interpolation strength of the $k$-th attribute $z^{k}$. The $k$-th attribute changes from sample $i$ to $j$ accordingly as $\boldsymbol{v}^{k}$ alters from $0$ to $1$. Thus, interpolation in the latent feature space should correspond to interpolation in the attribute space, if all possible values of $\boldsymbol{z}$ form an attribute space. 
The relation between the latent space and the attribute space initially defined in Equation~\ref{eqn:attribute_rep_1} can be re-formulated as
\begin{equation}
\label{eqn:attribute_rep_2}
\begin{aligned}
\mathcal{I}\left(\mathcal{C}_{\boldsymbol{v}}\left(F_{i}, F_{j}\right)\right)=\mathcal{C}_{\boldsymbol{v}}^{\prime}\left(\mathcal{I}\left(F_{i}\right), \mathcal{I}\left(F_{j}\right)\right), \forall \, \boldsymbol{v} \in[0,1]^{(c+1) \times 1}.
\end{aligned}
\end{equation}
In Equation~\ref{eqn:attribute_rep_2}, $\mathcal{C}_{\boldsymbol{v}}^{\prime}(\boldsymbol{z}_{i},\boldsymbol{z}_{j})$ can be viewed as an interpolation function defined in the attribute space that is controlled by $\boldsymbol{v}$. 
$\mathcal{C}_{\boldsymbol{v}}^{\prime}\left(\boldsymbol{z}_{i}, \boldsymbol{z}_{j}\right)$ is defined as
\begin{equation}
\label{eqn:interp_in_attribute_space}
\begin{aligned}
\mathcal{C}_{\boldsymbol{v}}^{\prime}\left(\boldsymbol{z}_{i}, \boldsymbol{z}_{j}\right)=\left[\mathcal {C }_{\boldsymbol{v}}^{\prime}\left(\boldsymbol{z}_{i}, \boldsymbol{z}_{j}\right)^{1} \cdots, \mathcal{C}_{\boldsymbol{v}}^{\prime}\left(\boldsymbol{z}_{i}, \boldsymbol{z}_{j}\right)^{c+1}\right],
\end{aligned}
\end{equation}
where $\mathcal{C}_{v}^{\prime}\left(\boldsymbol{z}_{i}, \boldsymbol{z}_{j}\right)^{k}=\boldsymbol{z}_{i}^{k}+\boldsymbol{v}^{k}\left(\boldsymbol{z}_{j}^{k}-\boldsymbol{z}_{i}^{k}\right)$.
Each dimension $\boldsymbol{v}^c$ sets the interpolation strength of each attribute between the two samples. 
There is a critical assumption underlying in Equation~\ref{eqn:attribute_rep_2}, which is the equivalence of the structure in the latent space and the attribute space defined by operations $\mathcal{C}_{\boldsymbol{v}}(\cdot)$ and $\mathcal{C}_{\boldsymbol{v}}^{\prime}(\cdot)$, \textit{i.e.}, \textit{isomorphism}.
The left-hand side of Equation~\ref{eqn:attribute_rep_2} means the attribute values of morphing samples $\mathcal{C}_{\boldsymbol{v}}(F_{j}-F_{i})$, and the right side represents attribute values of the two samples. They are expected to be equal since both sides are controlled by the same control vector $\boldsymbol{v}$. 
To ensure this, we train a network to approximate $\mathcal{C}(\cdot)$ and 
establish a one-to-one mapping from a unique identity element in the latent space to the corresponding one in the attribute space. Specifically, we train the attribute classification network $\mathcal{I}^{\prime}(\cdot)$ to map the interpolated feature $\mathcal{C}_{\boldsymbol{v}}(F_{j}-F_{i})=F_{i}+\sum_{k=1}^{c+1} \boldsymbol{v}^{k} \mathcal{T}^{k}(F_{j}-F_{i})$ to attribute $z_i$.
This can be achieved by minimizing the cross-entropy loss
\begin{equation}
\label{eqn:isomp_loss}
\begin{aligned}
\min_{\boldsymbol{\mathcal{C}}_{\boldsymbol{v}}} \mathcal{L}_{\boldsymbol{\mathcal{C}}_{isp}}=\mathbb{E}[-\mathcal{C}_{\boldsymbol{v}}^{\prime}(z_{i}, z_{j}) \log (\mathcal{I}^{\prime}(\mathcal{C}_{\boldsymbol{v}}(F_{i}, F_{j})))].
\end{aligned}
\end{equation}
During training, we assign uniformly random values to $\boldsymbol{v}$ to cover the whole feasible set.

Besides, similar to the perceptual losses~\cite{johnson2016perceptual}, we introduce a loss defined on feature space, which is formulated as
\begin{equation}
\label{eqn:feat_loss}
\begin{aligned}
\mathcal{L}_{\mathcal{C}_{t}}=\left\|{\mathcal{C}_{\boldsymbol{v}}}\left(F_{i}, F_{j}\right)-F_{j}\right\|^{2}. %
\end{aligned}
\end{equation}

To summarize, the overall loss function $\mathcal{L}_{\boldsymbol{\mathcal{C}}}$ of Controller $\boldsymbol{\mathcal{C}}$ is
\begin{equation}
\begin{aligned}
\mathcal{L}_{\boldsymbol{\mathcal{C}}}=\lambda_{GAN_{\boldsymbol{\mathcal{C}}}} \mathcal{L}_{GAN_{\boldsymbol{E}, \boldsymbol{\mathcal{C}}}}+\lambda_{{{\boldsymbol{\mathcal{C}}}_{isp}}} \mathcal{L}_{{\boldsymbol{\mathcal{C}}}_{isp}}+\lambda_{{\boldsymbol{\mathcal{C}}}_{t}} \mathcal{L}_{{\boldsymbol{\mathcal{C}}}_{t}}
\end{aligned}
\end{equation}
where $\mathcal{L}_{GAN_{\boldsymbol{E}, \boldsymbol{\mathcal{C}}}}$, $\mathcal{L}_{{\boldsymbol{\mathcal{C}}}_{isp}}$ and $\mathcal{L}_{{\boldsymbol{\mathcal{C}}}_{t}}$ are defined in Equation~\ref{eqn:adv_loss}, Equation~\ref{eqn:isomp_loss} and Equation~\ref{eqn:feat_loss}, respectively, and $\lambda_{GAN_{\boldsymbol{\mathcal{C}}}}$, $\lambda_{{\boldsymbol{\mathcal{C}}}_{isp}}$ and $\lambda_{{\boldsymbol{\mathcal{C}}}_{t}}$ are all set to $1$ in our experiments.

\begin{figure}[t]
\noindent\begin{minipage}{0.5\textwidth}
\includegraphics[width=1\linewidth]{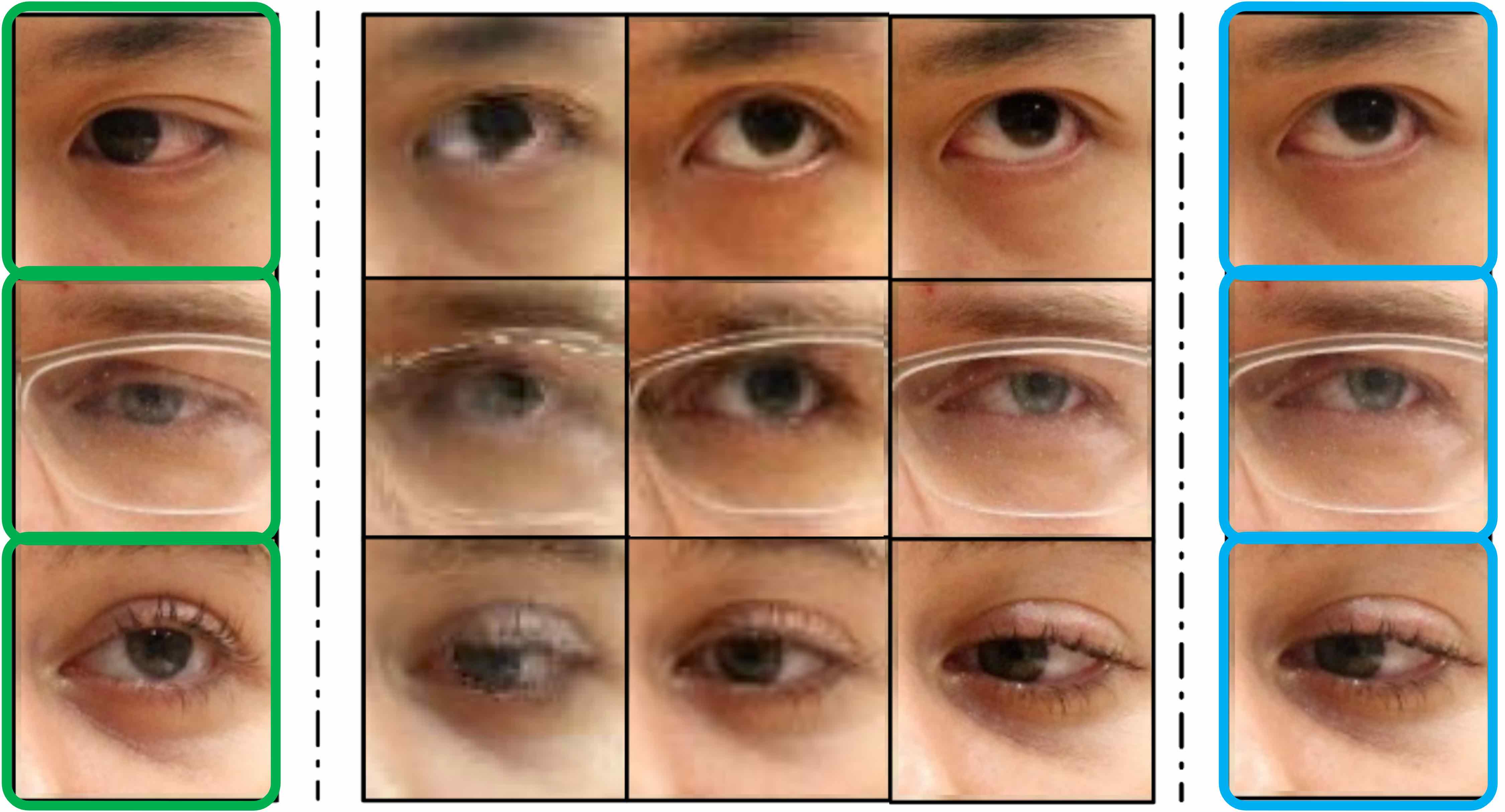}
\end{minipage}
\smallskip
\noindent\begin{minipage}{0.10\textwidth}
\centering
\scriptsize{Source}
\end{minipage}
\noindent\begin{minipage}{0.10\textwidth}
\flushright
\scriptsize{DeepWarp~\cite{ganin2016deepwarp}}
\end{minipage}
\noindent\begin{minipage}{0.09\textwidth}
\centering
\scriptsize{He~\textit{et al.}~\cite{he2019gaze}}
\end{minipage}
\noindent\begin{minipage}{0.075\textwidth}
\centering
\scriptsize{Ours}
\end{minipage}
\noindent\begin{minipage}{0.095\textwidth}
\flushright
\scriptsize{Ground-Truth}
\end{minipage}
\caption{\textbf{Gaze redirection comparison.} 
Detailed discussion in Section~\ref{sec:redirection_results}.}
\label{fig:redirection}
\end{figure}

\begin{table}[t]
\caption{Attributes and corresponding labels. The $1$st-$3$rd columns: dimension index in the control vector, name of attributes, corresponding attribute labels.}
\label{tab:attribute}
\begin{center}
\scalebox{0.95}{
\begin{tabular}{c|c|c}
\toprule
Dim  & Attribute & Labels  \\\hline
 1 &head pose &$0^\circ, \pm 15^\circ, \pm 30^\circ$ \\\hline
 2 &gaze yaw  & $0^\circ, \pm 5^\circ, \pm 10^\circ, \pm 15^\circ, \pm 25^\circ, \pm 35^\circ, \pm 45^\circ$ \\\hline
 3 & gaze pitch  & $0^\circ, \pm 10^\circ, \pm 15^\circ, \pm 25^\circ, \pm 35^\circ$  \\
\bottomrule
\end{tabular}}
\end{center}
\end{table}

\subsection{Control Mechanism}
\label{sec:control_mechanism}
Figure~\ref{fig:control_mechanism} shows the control mechanism for gaze redirection and interpolation. 
The abbreviations P, H, V and O are head pose (P), vertical gaze direction (pitch, V), horizontal gaze direction (yaw, H) and miscellaneous attributes, which are also consistent with the four branches in Figure~\ref{fig:overview}.

For gaze redirection, the inputs are a gaze patch of a certain person as the source image and that of another person (whose appearance may vary considerably) with the desired directions as the target.
For gaze interpolation, the inputs are two gaze patches of the same person. 
We can set the strength and order of the control vector to select features of each attribute and generate the desired results.
For example, 
given $(-15\text{P}, 0\text{V}, 15\text{H})$ as the source and $(15\text{P}, 0\text{V}, -15\text{H})$ as the target, we can set the control vector $\boldsymbol{v}_1$ as $[0.5,0.667,0,0]$ to generate a specific intermediate result $(0\text{P}, 0\text{V}, -5\text{H})$, as illustrated in Figure~\ref{fig:control_mechanism}(a). Similarly, we can generate predictable interpolation sequences in different \textbf{orders} using different control vectors. 
If we want to redirect the same source image to the desired direction $(0\text{P}, 0\text{V}, -5\text{H})$ defined in the reference image, we can set $\boldsymbol{v}_2$ as $[1.0,1.0,0,0]$, which alters the target attribute while keeping others almost intact, as illustrated in Figure~\ref{fig:control_mechanism}(b). 
The appearance of the reference image is allowed to be dramatically different from the source. This means that our method can be extended into \textbf{one-shot}~\cite{liu2019few} inference, using the synthetic image from UnityEyes~\cite{wood2016UnityEyes} or SynthesEyes~\cite{wood2015SynthesEyes}, which covers the full space of gaze direction, as the reference to produce an image with the unseen target direction.

The ``{\verb|O|}'' branch in Figure~\ref{fig:control_mechanism} represents other attributes like glass, eyebrow, skin color and hair as aforementioned. Both the last elements in $\boldsymbol{v}^4_1$ and $\boldsymbol{v}^4_2$ are set as $0$, which means that these secondary attributes remain unchanged during the generation. But we can also set $\boldsymbol{v}^4_1$ as any value in $[0,1]$ to generate \textbf{diverse} results, which makes our method multi-modal~\cite{zhu2017toward,drit,xia2019explicit}. On the one hand, the ``{\verb|O|}'' branches describe the common attributes of the same person when performing interpolation. On the other hand, the appearance of the two input images may be slightly different (not the attributes of the subject in the picture), which is mostly caused by the light and shade during photo capturing. This also indicates that our method can be extended to integrate more attributes like illumination and exposure.

Through the proposed feature disentanglement and control mechanism, we have unified two completely different tasks into the same framework, while avoiding the deficiencies in covering the gaze space of vector-based redirection methods.
Notice that we use relative offset of each attribute of two inputs in practice, the illustration in Figure~\ref{fig:control_mechanism} is simplified for easy understanding. 

\section{Dataset}
\label{sec:data}

\subsection{Why Do We Need A New Gaze Dataset?}
Unfolding the natural image manifold to a flat and smooth latent space is the key for our method to perform precise redirection and continuous interpolation, including the more challenging one-shot and extrapolation.
In addition to designing the network structure to allow the encoder to obtain a smooth and flat latent space, a dataset with a relatively larger range of directions is also crucial for the model to learn such latent space. Furthermore, such a high-quality gaze dataset can also benefit other researchers in related areas.

\subsection{Dataset Collection}

\begin{figure}
\begin{center}
\includegraphics[width=.92\linewidth]{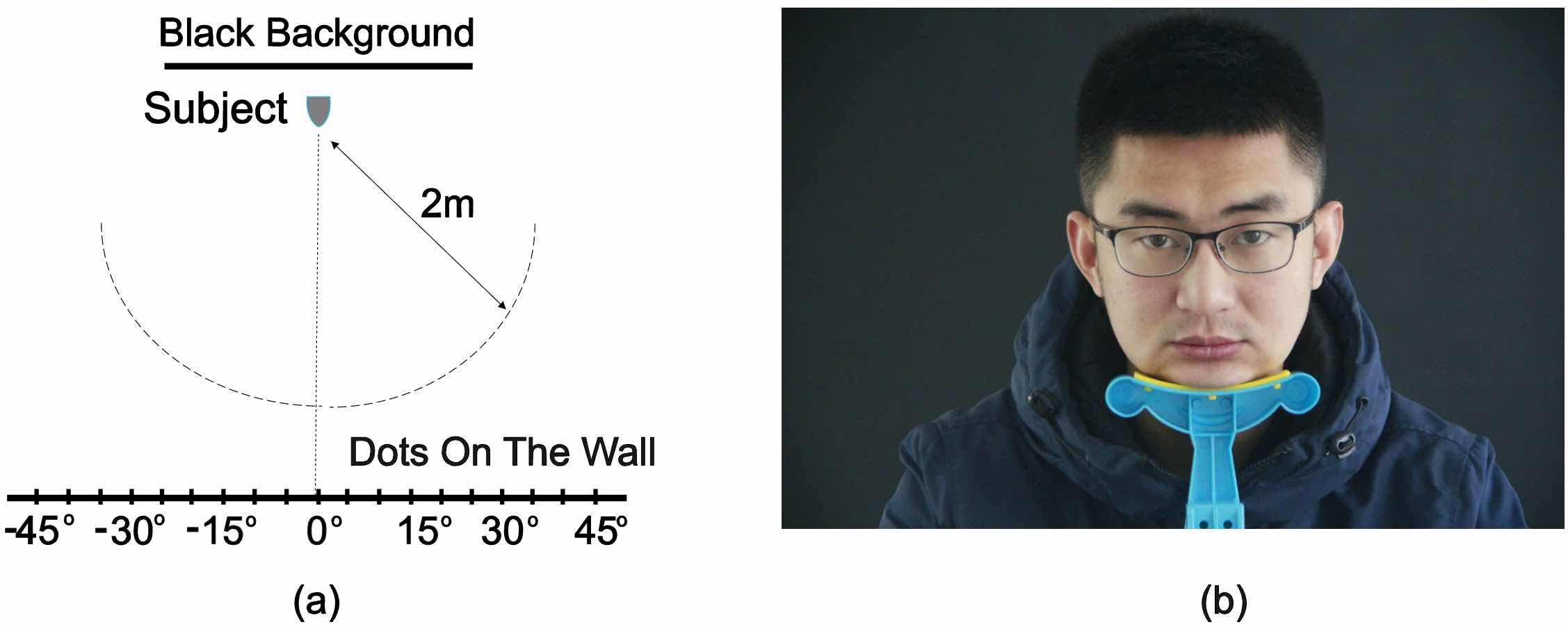}
\end{center}
\caption{\textbf{Data Collection Setup (a) and an example (b). For more details, please see Section~\ref{sec:data}.}}
\label{fig:collection}
\end{figure}

We collect images in the same way as~\cite{smith2013gaze}.
As shown in Figure~\ref{fig:collection}, subjects were seated in a fixed location in front of a black background, and a $13 \times 9$ grid of dots was attached to a wall in front of them. The dots were placed by the angle-distance relation. For each subject and head pose, we took images of the subject gazing at each dot of the pose's corresponding grid of dots.
Then, we collect a high-quality gaze dataset that contains $29,250$ images of $50$ subjects with the resolution of $5,184\times3,456$. We use resized eye patch images following the same procedure as in~\cite{he2019gaze} for our experiments. 
We have three gaze-related attributes with certain labels, \textit{i.e.}, head pose, gaze yaw and pitch direction, as shown in Table~\ref{tab:attribute}.
We randomly select 85 subjects from the Columbia Gaze~\cite{smith2013gaze} and ours for training, and use the others for testing. For fair comparison on gaze redirection, we also train our model only with Columbia Gaze ~\cite{smith2013gaze}.

\begin{table*}[htbp]
\caption{Comparison of some popular gaze datasets.}
\label{tab:gaze_data}
\begin{center}
\begin{tabular}{cccccccc}
\toprule
Dataset & Real& High Res& Constrained & Annotation Type & Num. Image &Head Pose &Gaze Range\\
\hline\hline
CMU Multi-Pie~\cite{Gross2010Multi}&\ncmark&\nxmark&\ncmark&Facial landmarks&755,370&\ncmark&Small\\
Gaze Capture~\cite{krafka2016eye} &\ncmark &\nxmark &\nxmark &2D position on screen &$\textgreater$2.5M &\nxmark &Small\\
SynthesEyes~\cite{wood2015SynthesEyes} &\nxmark &\ncmark &\nxmark &Gaze vector & 11,382 &\ncmark & Full\\
UnityEyes~\cite{wood2016UnityEyes} &\nxmark &\ncmark &\nxmark &Gaze vector &1,000,000 &\ncmark&Full\\
MPII Gaze~\cite{zhang15_cvpr,Zhang2017gaze} &\ncmark &\nxmark &\nxmark &Gaze vector &213,659 &\ncmark &Small\\
UT Multi-View~\cite{Sugano2014Learning}&\ncmark &\nxmark &\ncmark &Gaze vector&1,152,000 &\ncmark &Large\\
Columbia~\cite{smith2013gaze} &\ncmark &\ncmark &\ncmark &Gaze vector& 5,880 &\ncmark &Medium\\
\hline\hline
Ours&\ncmark &\ncmark &\ncmark &Gaze vector & 29,250 &\ncmark &Large\\
\bottomrule
\end{tabular}
\end{center}
\end{table*}

\subsection{Dataset Analysis}
There are several important properties that a high-quality gaze redirection dataset should own: realness, high-resolution, collected in the constrained environment, with precise annotation and a large range of directions.
We compare a range of RGB gaze datasets in Table~\ref{tab:gaze_data}.
Some publicly available gaze datasets, such as MPIIGaze~\cite{zhang15_cvpr}, Gaze Capture~\cite{krafka2016eye}, UT Multi-View~\cite{Sugano2014Learning} or CMU Multi-Pie~\cite{Gross2010Multi}, only provide low-resolution images and would therefore introduce an unwanted bias towards low-quality
images. 
Those~\cite{krafka2016eye,zhang15_cvpr} proposed for gaze estimation are collected under unconstrained environmental conditions and vary in background and illumination, which means these datasets are not suitable for high-quality gaze redirection.
Some recent learning-based gaze redirection studies either learn from synthetic data~\cite{odobez2019gaze} (\textit{e.g.}, SynthesEyes~\cite{wood2015SynthesEyes}, UnityEyes~\cite{wood2016UnityEyes}), or use limited high-quality dataset~\cite{he2019gaze} (\textit{e.g.}, Columbia Gaze~\cite{smith2013gaze}).
The synthetic data~\textit{e.g.}, SynthesEyes~\cite{wood2015SynthesEyes} and UnityEyes~\cite{wood2016UnityEyes}, covers the full space of gaze direction but methods trained on them are not as photo-realistic as those trained on high-quality real images.
The Columbia Gaze dataset~\cite{smith2013gaze} is a high-resolution, publicly available human gaze dataset collected from 56 subjects in a constrained environment.
The head poses of each subject are discrete values in the set [$0^\circ, \pm 15^\circ, \pm 30^\circ$]. 
For each head pose, there are 21 gaze directions, which are the combinations of three pitch angles [$0^\circ, \pm 10^\circ$], and seven yaw angles [$0^\circ, \pm 5^\circ, \pm 10^\circ, \pm 15^\circ$]. However, it does not cover an enough range of gaze direction. 
According to the Driver and Vehicle Licensing Agency (DVLA) in the United Kingdom, the normal range of the human eye is about $120^\circ$ of horizontal field and $40^\circ$ of vertical field. In our collection procedure, we found that it is about $45^\circ$ horizontally and $35^\circ$ vertically if focused accurately on a dot. Out of this range, the obtained gaze images would look like almost the same in a certain direction. 

\section{Experiments}
\subsection{Metrics}
It remains an open problem to effectively evaluate the quality of generated images in image generation tasks. Gaze redirection models are required to be precise in redirecting and to produce photo-realistic and consistent images. Correspondingly, the evaluation metrics need to be able to assess these aspects.
Similarly to He \textit{et al.}~\cite{he2019gaze}, we use the mean squared error (MSE), Learned Perceptual Image Patch Similarity (LPIPS)~\cite{zhang2018perceptual} and Gaze Estimation Error~\cite{park2018deep}, as the metrics to measure the similarity, perception quality and gaze direction error, respectively.

\subsection{Gaze Redirection}
\label{sec:redirection_results}

\vspace*{2mm}\noindent {\bf Baseline Model.}
We adopt DeepWarp~\cite{ganin2016deepwarp} and a recent GAN-based method~\cite{he2019gaze} as our baselines. We did not compare with recent GazeAdaptation~\cite{yu2019improving} and GazeDirector~\cite{wood2018gazedirector} since their implementations are not publicly available.

\vspace*{2mm}\noindent {\bf Qualitative Evaluation.}
Figure~\ref{fig:redirection} shows the generated gaze images examples. The reference image for certain gaze direction is randomly selected.
Although both DeepWarp~\cite{ganin2016deepwarp} and He \textit{et al.}~\cite{he2019gaze} are capable of redirecting the gaze, the generated images have several obvious defects. For example, DeepWarp~\cite{ganin2016deepwarp} prodeces blurry textures and boundaries such as skin and eyebrows, and the shapes of certain areas, such as the edges of eyelid, iris and eyeglasses, are distorted; the generated gaze images of He \textit{et al.}~\cite{he2019gaze} are more faithful to the input images, but suffer from unnatural pupil when redirecting to extreme angles as shown in the last row of Figure~\ref{fig:redirection}. 
In contrast, our method achieves better performance on visual plausibility and redirection precision.

\vspace*{2mm}\noindent {\bf Quantitative Evaluation.}
Quantitative evaluation is summarized in Table~\ref{tab:quantitative_evaluation}.
We can see that our method achieves the lowest LPIPS score~\cite{zhang2018perceptual} at every correction angle, which indicates that our method can generate gaze images that are perceptually more similar to the ground-truth images. This observation is consistent with the qualitative evaluation.

\vspace*{2mm}\noindent {\bf User Study.} We ask $21$ users to pick the gaze image that looks more realistic than the other.
The generated gaze images are split into three groups as in~\cite{he2019gaze}, $[4.9^{\circ}$, $15.0^{\circ}]$, $(15.0^{\circ}$, $25.0^{\circ}]$, $(25.0^{\circ}$, $35.9^{\circ}]$. In each group, we randomly choose 19 images generated by both methods with the same input image and gaze direction. For ours, we randomly choose a gaze image with the desired direction as the reference image. Three images are displayed alongside and shuffled randomly.
The results are shown in Table~\ref{tab:user_study}.
 
\subsection{Gaze Interpolation}
\label{sec:interp_results}
The results of continuous gaze interpolation is shown in Figure~\ref{fig:interpolation}, where $x_s$ (green rectangles) and $x_t$ (blue rectangles) are two gaze images randomly sampled from the same person, and moving from source image $x_s$ toward target image $x_t$ in the latent space  gradually produces continuous realistic images $x_g$. 
Given different control vectors $\boldsymbol{v}$, we can select the strength of intermediate results. For example, the first and third rows change only gaze directions with constant head pose, while the second and forth rows perform the opposite. It can be seen that other attributes like eyebrow, glass, hair and skin color are well-preserved in the redirected gaze images, which means that our model works consistently well in generating person-specific gaze images. We also gradually increase the strength of the last dimension of $\boldsymbol{v}$ to produce multi-modal results, as shown in the third row.
Furthermore, 
our method can also perform extrapolation, as demonstrated in the last column of Figure~\ref{fig:interpolation}.
This indicates that the encoder has unfolded the natural image manifold, leading to a well-learned flat and smooth latent space that allows precise control including redirection, interpolation and extrapolation.

\begin{table}[th]
\begin{center}
\caption{Comparison with the state-of-the-arts in terms of the MSE, LPIPS and Gaze Estimation Error of different direction groups. $\downarrow$ means the lower the better.}
\label{tab:quantitative_evaluation}
\begin{tabular}{cccc}
\toprule
Method& MSE $\downarrow$ & LPIPS$\downarrow$ &Gaze Error$\downarrow$\\
\midrule
Deepwarp~\cite{ganin2016deepwarp} &126.71&0.083 &$15.3^{\circ}$\\
He \textit{et al.}~\cite{he2019gaze} &72.15&0.066 &$8.7^{\circ}$\\
Ours &\textbf{56.90} &\textbf{0.037} &$\mathbf{6.3^{\circ}}$\\
\bottomrule
\end{tabular}
\end{center}
\end{table}

\begin{table}[th]
\caption{Voting results from the user study comparing DeepWarp~\cite{ganin2016deepwarp} and He \textit{et al.}~\cite{he2019gaze} with our method. Each row sums up to 100$\%$.}
\label{tab:user_study}
\begin{center}
\begin{tabular}{cccc}
\hline Group & DeepWarp~\cite{ganin2016deepwarp} &He \textit{et al.}~\cite{he2019gaze}& Ours \\
\hline \hline
$[4.9^{\circ}, 15.0^{\circ}]$ & $4.7\%$ &$28.6\%$& $\mathbf{66.7}\%$\\
$(15.0^{\circ}, 25.0^{\circ}]$ & $14.3\%$ &$33.3\%$ &$\mathbf{52.4\%}$ \\
$(3.2^{\circ}, 29.8^{\circ}]$ & $9.5\%$ &$42.9\%$& $\mathbf{47.6\%}$ \\
\hline
\end{tabular}
\end{center}
\end{table}

\section{Conclusion}
In this paper, we have introduced \textit{interpGaze},
a framework for controllable gaze redirection that can easily achieve both precise redirection and continuous interpolation, and even extrapolation by altering a control vector.
Through the proposed feature disentanglement and control mechanism, we have unified two completely different tasks into the same framework, while avoiding the deficiencies in covering the gaze space of vector-based redirection methods.
With our elaborate network design and our new dataset of large range gaze directions, our method can cover a large space of directions for gaze redirection and interpolation. 

\section{Acknowledgement}
This research was partially supported by the Key Program of National Natural Science Foundation of China under Grant No. U1903213, the Dedicated Fund for Promoting High-Quality Economic Development in Guangdong Province (Marine Economic Development Project: GDOE[2019]A45), and the Shenzhen Science and Technology Project under Grant (ZDYBH201900000002, JCYJ20180508152042002).
\bibliographystyle{ACM-Reference-Format}
\bibliography{acmart}


\begin{thebibliography}{47}

%%% the Web address.

\ifx \showCODEN    \undefined \def \showCODEN     #1{\unskip}     \fi
\ifx \showDOI      \undefined \def \showDOI       #1{#1}\fi
\ifx \showISBNx    \undefined \def \showISBNx     #1{\unskip}     \fi
\ifx \showISBNxiii \undefined \def \showISBNxiii  #1{\unskip}     \fi
\ifx \showISSN     \undefined \def \showISSN      #1{\unskip}     \fi
\ifx \showLCCN     \undefined \def \showLCCN      #1{\unskip}     \fi
\ifx \shownote     \undefined \def \shownote      #1{#1}          \fi
\ifx \showarticletitle \undefined \def \showarticletitle #1{#1}   \fi
\ifx \showURL      \undefined \def \showURL       {\relax}        \fi
% invisible to TeX
\providecommand\bibfield[2]{#2}
\providecommand\bibinfo[2]{#2}
\providecommand\natexlab[1]{#1}
\providecommand\showeprint[2][]{arXiv:#2}

\bibitem[\protect\citeauthoryear{Abdal, Qin, and Wonka}{Abdal
  et~al\mbox{.}}{2019}]%
        {abdal2019embed}
\bibfield{author}{\bibinfo{person}{Rameen Abdal}, \bibinfo{person}{Yipeng Qin},
  {and} \bibinfo{person}{Peter Wonka}.} \bibinfo{year}{2019}\natexlab{}.
\newblock \showarticletitle{Image2StyleGAN: How to Embed Images Into the
  StyleGAN Latent Space?}
\newblock \bibinfo{journal}{\emph{International Conference on Computer Vision}}
  (\bibinfo{year}{2019}).
\newblock


\bibitem[\protect\citeauthoryear{Banf and Blanz}{Banf and Blanz}{2009}]%
        {banf2009example}
\bibfield{author}{\bibinfo{person}{Michael Banf} {and} \bibinfo{person}{Volker
  Blanz}.} \bibinfo{year}{2009}\natexlab{}.
\newblock \showarticletitle{Example-based rendering of eye movements}. In
  \bibinfo{booktitle}{\emph{Computer Graphics Forum}},
  Vol.~\bibinfo{volume}{28}. Wiley Online Library, \bibinfo{pages}{659--666}.
\newblock


\bibitem[\protect\citeauthoryear{Bengio, Mesnil, Dauphin, and Rifai}{Bengio
  et~al\mbox{.}}{2013}]%
        {bengio2013distill}
\bibfield{author}{\bibinfo{person}{Yoshua Bengio},
  \bibinfo{person}{Gr{\'{e}}goire Mesnil}, \bibinfo{person}{Yann~N. Dauphin},
  {and} \bibinfo{person}{Salah Rifai}.} \bibinfo{year}{2013}\natexlab{}.
\newblock \showarticletitle{Better Mixing via Deep Representations}. In
  \bibinfo{booktitle}{\emph{Proceedings of the 30th International Conference on
  Machine Learning, {ICML} 2013, Atlanta, GA, USA, 16-21 June 2013}}.
  \bibinfo{pages}{552--560}.
\newblock


\bibitem[\protect\citeauthoryear{Berthelot, Raffel, Roy, and
  Goodfellow}{Berthelot et~al\mbox{.}}{2019}]%
        {Berthelot19RRG}
\bibfield{author}{\bibinfo{person}{David Berthelot}, \bibinfo{person}{Colin
  Raffel}, \bibinfo{person}{Aurko Roy}, {and} \bibinfo{person}{Ian~J.
  Goodfellow}.} \bibinfo{year}{2019}\natexlab{}.
\newblock \showarticletitle{Understanding and Improving Interpolation in
  Autoencoders via an Adversarial Regularizer}. In
  \bibinfo{booktitle}{\emph{Proceedings of International Conference on Learning
  Representations}}.
\newblock


\bibitem[\protect\citeauthoryear{Chen, Song, and Hilliges}{Chen
  et~al\mbox{.}}{2019a}]%
        {chen2019mono}
\bibfield{author}{\bibinfo{person}{Xu Chen}, \bibinfo{person}{Jie Song}, {and}
  \bibinfo{person}{Otmar Hilliges}.} \bibinfo{year}{2019}\natexlab{a}.
\newblock \showarticletitle{Monocular Neural Image Based Rendering with
  Continuous View Control}. In \bibinfo{booktitle}{\emph{International
  Conference on Computer Vision}}.
\newblock


\bibitem[\protect\citeauthoryear{Chen, Lin, Shu, Li, Tao, Shen, Ye, and
  Jia}{Chen et~al\mbox{.}}{2018}]%
        {Chen18FFP}
\bibfield{author}{\bibinfo{person}{Ying{-}Cong Chen}, \bibinfo{person}{Huaijia
  Lin}, \bibinfo{person}{Michelle Shu}, \bibinfo{person}{Ruiyu Li},
  \bibinfo{person}{Xin Tao}, \bibinfo{person}{Xiaoyong Shen},
  \bibinfo{person}{Yangang Ye}, {and} \bibinfo{person}{Jiaya Jia}.}
  \bibinfo{year}{2018}\natexlab{}.
\newblock \showarticletitle{Facelet-Bank for Fast Portrait Manipulation}. In
  \bibinfo{booktitle}{\emph{Proceedings of International Conference on Computer
  Vision and Pattern Recognition}}. \bibinfo{pages}{3541--3549}.
\newblock


\bibitem[\protect\citeauthoryear{Chen, Xu, Tian, and Jia}{Chen
  et~al\mbox{.}}{2019b}]%
        {chen2019Homomorphic}
\bibfield{author}{\bibinfo{person}{Ying-Cong Chen}, \bibinfo{person}{Xiaogang
  Xu}, \bibinfo{person}{Zhuotao Tian}, {and} \bibinfo{person}{Jiaya Jia}.}
  \bibinfo{year}{2019}\natexlab{b}.
\newblock \showarticletitle{Homomorphic Latent Space Interpolation for Unpaired
  Image-to-image Translation}. In \bibinfo{booktitle}{\emph{CVPR}}.
\newblock


\bibitem[\protect\citeauthoryear{Choi, Choi, Kim, Ha, Kim, and Choo}{Choi
  et~al\mbox{.}}{2018}]%
        {choi2018stargan}
\bibfield{author}{\bibinfo{person}{Yunjey Choi}, \bibinfo{person}{Minje Choi},
  \bibinfo{person}{Munyoung Kim}, \bibinfo{person}{Jung-Woo Ha},
  \bibinfo{person}{Sunghun Kim}, {and} \bibinfo{person}{Jaegul Choo}.}
  \bibinfo{year}{2018}\natexlab{}.
\newblock \showarticletitle{Stargan: Unified generative adversarial networks
  for multi-domain image-to-image translation}. In
  \bibinfo{booktitle}{\emph{Proceedings of the IEEE Conference on Computer
  Vision and Pattern Recognition}}. \bibinfo{pages}{8789--8797}.
\newblock


\bibitem[\protect\citeauthoryear{Criminisi, Shotton, Blake, and Torr}{Criminisi
  et~al\mbox{.}}{2003}]%
        {criminisi2003gaze}
\bibfield{author}{\bibinfo{person}{Antonio Criminisi}, \bibinfo{person}{Jamie
  Shotton}, \bibinfo{person}{Andrew Blake}, {and} \bibinfo{person}{Philip~HS
  Torr}.} \bibinfo{year}{2003}\natexlab{}.
\newblock \showarticletitle{Gaze Manipulation for One-to-one
  Teleconferencing.}. In \bibinfo{booktitle}{\emph{ICCV}},
  Vol.~\bibinfo{volume}{3}. \bibinfo{pages}{13--16}.
\newblock


\bibitem[\protect\citeauthoryear{Ganin, Kononenko, Sungatullina, and
  Lempitsky}{Ganin et~al\mbox{.}}{2016}]%
        {ganin2016deepwarp}
\bibfield{author}{\bibinfo{person}{Yaroslav Ganin}, \bibinfo{person}{Daniil
  Kononenko}, \bibinfo{person}{Diana Sungatullina}, {and}
  \bibinfo{person}{Victor Lempitsky}.} \bibinfo{year}{2016}\natexlab{}.
\newblock \showarticletitle{Deepwarp: Photorealistic image resynthesis for gaze
  manipulation}. In \bibinfo{booktitle}{\emph{European Conference on Computer
  Vision}}. Springer, \bibinfo{pages}{311--326}.
\newblock


\bibitem[\protect\citeauthoryear{Gross, Matthews, Cohn, Kanade, and
  Baker}{Gross et~al\mbox{.}}{2010}]%
        {Gross2010Multi}
\bibfield{author}{\bibinfo{person}{R Gross}, \bibinfo{person}{I Matthews},
  \bibinfo{person}{J Cohn}, \bibinfo{person}{T Kanade}, {and}
  \bibinfo{person}{S Baker}.} \bibinfo{year}{2010}\natexlab{}.
\newblock \showarticletitle{Multi-PIE}. In \bibinfo{booktitle}{\emph{Image and
  Vision Computing}}. \bibinfo{pages}{807--813}.
\newblock


\bibitem[\protect\citeauthoryear{Gulrajani, Ahmed, Arjovsky, Dumoulin, and
  Courville}{Gulrajani et~al\mbox{.}}{2017}]%
        {gulrajani2017improved}
\bibfield{author}{\bibinfo{person}{Ishaan Gulrajani}, \bibinfo{person}{Faruk
  Ahmed}, \bibinfo{person}{Martin Arjovsky}, \bibinfo{person}{Vincent
  Dumoulin}, {and} \bibinfo{person}{Aaron~C Courville}.}
  \bibinfo{year}{2017}\natexlab{}.
\newblock \showarticletitle{Improved training of wasserstein gans}. In
  \bibinfo{booktitle}{\emph{Advances in Neural Information Processing
  Systems}}. \bibinfo{pages}{5767--5777}.
\newblock


\bibitem[\protect\citeauthoryear{He, Spurr, Zhang, and Hilliges}{He
  et~al\mbox{.}}{2019}]%
        {he2019gaze}
\bibfield{author}{\bibinfo{person}{Zhe He}, \bibinfo{person}{Adrian Spurr},
  \bibinfo{person}{Xucong Zhang}, {and} \bibinfo{person}{Otmar Hilliges}.}
  \bibinfo{year}{2019}\natexlab{}.
\newblock \showarticletitle{Photo-realistic Monocular Gaze Redirection using
  Generative Adversarial Networks}.
\newblock \bibinfo{journal}{\emph{International Conference on Computer Vision}}
  (\bibinfo{year}{2019}).
\newblock


\bibitem[\protect\citeauthoryear{Iizuka, Simo{-}Serra, and Ishikawa}{Iizuka
  et~al\mbox{.}}{2017}]%
        {Iizuka2017completion}
\bibfield{author}{\bibinfo{person}{Satoshi Iizuka}, \bibinfo{person}{Edgar
  Simo{-}Serra}, {and} \bibinfo{person}{Hiroshi Ishikawa}.}
  \bibinfo{year}{2017}\natexlab{}.
\newblock \showarticletitle{Globally and locally consistent image completion}.
\newblock \bibinfo{journal}{\emph{{ACM} Trans. Graph.}} \bibinfo{volume}{36},
  \bibinfo{number}{4} (\bibinfo{year}{2017}), \bibinfo{pages}{107:1--107:14}.
\newblock


\bibitem[\protect\citeauthoryear{Johnson, Alahi, and Fei-Fei}{Johnson
  et~al\mbox{.}}{2016}]%
        {johnson2016perceptual}
\bibfield{author}{\bibinfo{person}{Justin Johnson}, \bibinfo{person}{Alexandre
  Alahi}, {and} \bibinfo{person}{Li Fei-Fei}.} \bibinfo{year}{2016}\natexlab{}.
\newblock \showarticletitle{Perceptual losses for real-time style transfer and
  super-resolution}. In \bibinfo{booktitle}{\emph{European conference on
  computer vision}}. Springer, \bibinfo{pages}{694--711}.
\newblock


\bibitem[\protect\citeauthoryear{Karras, Laine, and Aila}{Karras
  et~al\mbox{.}}{2019}]%
        {kerras2019stylegan}
\bibfield{author}{\bibinfo{person}{Tero Karras}, \bibinfo{person}{Samuli
  Laine}, {and} \bibinfo{person}{Timo Aila}.} \bibinfo{year}{2019}\natexlab{}.
\newblock \showarticletitle{A Style-Based Generator Architecture for Generative
  Adversarial Networks}. In \bibinfo{booktitle}{\emph{CVPR}}.
  \bibinfo{pages}{4401--4410}.
\newblock


\bibitem[\protect\citeauthoryear{Krafka, Khosla, Kellnhofer, Kannan,
  Bhandarkar, Matusik, and Torralba}{Krafka et~al\mbox{.}}{2016}]%
        {krafka2016eye}
\bibfield{author}{\bibinfo{person}{Kyle Krafka}, \bibinfo{person}{Aditya
  Khosla}, \bibinfo{person}{Petr Kellnhofer}, \bibinfo{person}{Harini Kannan},
  \bibinfo{person}{Suchendra Bhandarkar}, \bibinfo{person}{Wojciech Matusik},
  {and} \bibinfo{person}{Antonio Torralba}.} \bibinfo{year}{2016}\natexlab{}.
\newblock \showarticletitle{Eye tracking for everyone}. In
  \bibinfo{booktitle}{\emph{CVPR}}. \bibinfo{pages}{2176--2184}.
\newblock


\bibitem[\protect\citeauthoryear{Krizhevsky, Sutskever, and Hinton}{Krizhevsky
  et~al\mbox{.}}{2012}]%
        {krizhevsky2012imagenet}
\bibfield{author}{\bibinfo{person}{Alex Krizhevsky}, \bibinfo{person}{Ilya
  Sutskever}, {and} \bibinfo{person}{Geoffrey~E Hinton}.}
  \bibinfo{year}{2012}\natexlab{}.
\newblock \showarticletitle{Imagenet classification with deep convolutional
  neural networks}. In \bibinfo{booktitle}{\emph{Advances in neural information
  processing systems}}. \bibinfo{pages}{1097--1105}.
\newblock


\bibitem[\protect\citeauthoryear{Kuster, Popa, Bazin, Gotsman, and
  Gross}{Kuster et~al\mbox{.}}{2012}]%
        {kuster2012gaze}
\bibfield{author}{\bibinfo{person}{Claudia Kuster}, \bibinfo{person}{Tiberiu
  Popa}, \bibinfo{person}{Jean-Charles Bazin}, \bibinfo{person}{Craig Gotsman},
  {and} \bibinfo{person}{Markus Gross}.} \bibinfo{year}{2012}\natexlab{}.
\newblock \showarticletitle{Gaze correction for home video conferencing}.
\newblock \bibinfo{journal}{\emph{ACM Transactions on Graphics (TOG)}}
  \bibinfo{volume}{31}, \bibinfo{number}{6} (\bibinfo{year}{2012}),
  \bibinfo{pages}{174}.
\newblock


\bibitem[\protect\citeauthoryear{Lee, Tseng, Huang, Singh, and Yang}{Lee
  et~al\mbox{.}}{2018}]%
        {drit}
\bibfield{author}{\bibinfo{person}{Hsin-Ying Lee}, \bibinfo{person}{Hung-Yu
  Tseng}, \bibinfo{person}{Jia-Bin Huang}, \bibinfo{person}{Maneesh~Kumar
  Singh}, {and} \bibinfo{person}{Ming-Hsuan Yang}.}
  \bibinfo{year}{2018}\natexlab{}.
\newblock \showarticletitle{Diverse Image-to-Image Translation via Disentangled
  Representations}. In \bibinfo{booktitle}{\emph{ECCV}}.
\newblock


\bibitem[\protect\citeauthoryear{Li, Lin, Ding, Liu, Zhu, and Han}{Li
  et~al\mbox{.}}{2020}]%
        {li2020gan}
\bibfield{author}{\bibinfo{person}{Muyang Li}, \bibinfo{person}{Ji Lin},
  \bibinfo{person}{Yaoyao Ding}, \bibinfo{person}{Zhijian Liu},
  \bibinfo{person}{Jun-Yan Zhu}, {and} \bibinfo{person}{Song Han}.}
  \bibinfo{year}{2020}\natexlab{}.
\newblock \showarticletitle{GAN Compression: Efficient Architectures for
  Interactive Conditional GANs}. In \bibinfo{booktitle}{\emph{Proceedings of
  the IEEE/CVF Conference on Computer Vision and Pattern Recognition}}.
\newblock


\bibitem[\protect\citeauthoryear{Lira, Merz, Ritchie, Cohen{-}Or, and
  Zhang}{Lira et~al\mbox{.}}{2020}]%
        {lira2020hop}
\bibfield{author}{\bibinfo{person}{Wallace~P. Lira}, \bibinfo{person}{Johannes
  Merz}, \bibinfo{person}{Daniel Ritchie}, \bibinfo{person}{Daniel Cohen{-}Or},
  {and} \bibinfo{person}{Hao~(Richard) Zhang}.}
  \bibinfo{year}{2020}\natexlab{}.
\newblock \showarticletitle{GANHopper: Multi-Hop {GAN} for Unsupervised
  Image-to-Image Translation}.
\newblock \bibinfo{journal}{\emph{CoRR}}  \bibinfo{volume}{abs/2002.10102}
  (\bibinfo{year}{2020}).
\newblock


\bibitem[\protect\citeauthoryear{Liu, Huang, Mallya, Karras, Aila, Lehtinen,
  and Kautz}{Liu et~al\mbox{.}}{2019}]%
        {liu2019few}
\bibfield{author}{\bibinfo{person}{Ming-Yu Liu}, \bibinfo{person}{Xun Huang},
  \bibinfo{person}{Arun Mallya}, \bibinfo{person}{Tero Karras},
  \bibinfo{person}{Timo Aila}, \bibinfo{person}{Jaakko Lehtinen}, {and}
  \bibinfo{person}{Jan Kautz}.} \bibinfo{year}{2019}\natexlab{}.
\newblock \showarticletitle{Few-Shot Unsupervised Image-to-Image Translation}.
  In \bibinfo{booktitle}{\emph{{IEEE} International Conference on Computer
  Vision (ICCV)}}.
\newblock


\bibitem[\protect\citeauthoryear{Mildenhall, Srinivasan, Tancik, Barron,
  Ramamoorthi, and Ng}{Mildenhall et~al\mbox{.}}{2020}]%
        {mildenhall2020scenes}
\bibfield{author}{\bibinfo{person}{Ben Mildenhall}, \bibinfo{person}{Pratul~P.
  Srinivasan}, \bibinfo{person}{Matthew Tancik}, \bibinfo{person}{Jonathan~T.
  Barron}, \bibinfo{person}{Ravi Ramamoorthi}, {and} \bibinfo{person}{Ren Ng}.}
  \bibinfo{year}{2020}\natexlab{}.
\newblock \showarticletitle{NeRF: Representing Scenes as Neural Radiance Fields
  for View Synthesis}. In \bibinfo{booktitle}{\emph{Proc. of the IEEE
  Conference on Computer Vision and Pattern Recognition (CVPR)}}.
\newblock


\bibitem[\protect\citeauthoryear{Olszewski, Tulyakov, Woodford, Li, and
  Luo}{Olszewski et~al\mbox{.}}{2019}]%
        {olszewski2019tbn}
\bibfield{author}{\bibinfo{person}{Kyle Olszewski}, \bibinfo{person}{Sergey
  Tulyakov}, \bibinfo{person}{Oliver Woodford}, \bibinfo{person}{Hao Li}, {and}
  \bibinfo{person}{Linjie Luo}.} \bibinfo{year}{2019}\natexlab{}.
\newblock \showarticletitle{Transformable Bottleneck Networks}. In
  \bibinfo{booktitle}{\emph{The IEEE International Conference on Computer
  Vision (ICCV)}}.
\newblock


\bibitem[\protect\citeauthoryear{Park, Spurr, and Hilliges}{Park
  et~al\mbox{.}}{2018}]%
        {park2018deep}
\bibfield{author}{\bibinfo{person}{Seonwook Park}, \bibinfo{person}{Adrian
  Spurr}, {and} \bibinfo{person}{Otmar Hilliges}.}
  \bibinfo{year}{2018}\natexlab{}.
\newblock \showarticletitle{Deep pictorial gaze estimation}. In
  \bibinfo{booktitle}{\emph{ECCV}}. \bibinfo{pages}{721--738}.
\newblock


\bibitem[\protect\citeauthoryear{Pumarola, Agudo, Martinez, Sanfeliu, and
  Moreno-Noguer}{Pumarola et~al\mbox{.}}{2019}]%
        {Pumarola_ijcv2019}
\bibfield{author}{\bibinfo{person}{A. Pumarola}, \bibinfo{person}{A. Agudo},
  \bibinfo{person}{A.M. Martinez}, \bibinfo{person}{A. Sanfeliu}, {and}
  \bibinfo{person}{F. Moreno-Noguer}.} \bibinfo{year}{2019}\natexlab{}.
\newblock \showarticletitle{GANimation: One-Shot Anatomically Consistent Facial
  Animation}.
\newblock  (\bibinfo{year}{2019}).
\newblock


\bibitem[\protect\citeauthoryear{Simonyan and Zisserman}{Simonyan and
  Zisserman}{2014}]%
        {simonyan2014very}
\bibfield{author}{\bibinfo{person}{Karen Simonyan} {and}
  \bibinfo{person}{Andrew Zisserman}.} \bibinfo{year}{2014}\natexlab{}.
\newblock \showarticletitle{Very deep convolutional networks for large-scale
  image recognition}.
\newblock \bibinfo{journal}{\emph{arXiv preprint arXiv:1409.1556}}
  (\bibinfo{year}{2014}).
\newblock


\bibitem[\protect\citeauthoryear{Smith, Yin, Feiner, and Nayar}{Smith
  et~al\mbox{.}}{2013}]%
        {smith2013gaze}
\bibfield{author}{\bibinfo{person}{Brian~A Smith}, \bibinfo{person}{Qi Yin},
  \bibinfo{person}{Steven~K Feiner}, {and} \bibinfo{person}{Shree~K Nayar}.}
  \bibinfo{year}{2013}\natexlab{}.
\newblock \showarticletitle{Gaze locking: passive eye contact detection for
  human-object interaction}. In \bibinfo{booktitle}{\emph{Proceedings of the
  26th annual ACM symposium on User interface software and technology}}. ACM,
  \bibinfo{pages}{271--280}.
\newblock


\bibitem[\protect\citeauthoryear{Sugano, Matsushita, and Sato}{Sugano
  et~al\mbox{.}}{2014}]%
        {Sugano2014Learning}
\bibfield{author}{\bibinfo{person}{Yusuke Sugano}, \bibinfo{person}{Yasuyuki
  Matsushita}, {and} \bibinfo{person}{Yoichi Sato}.}
  \bibinfo{year}{2014}\natexlab{}.
\newblock \showarticletitle{Learning-by-Synthesis for Appearance-Based 3D Gaze
  Estimation}. In \bibinfo{booktitle}{\emph{2014 IEEE Conference on Computer
  Vision and Pattern Recognition (CVPR)}}.
\newblock


\bibitem[\protect\citeauthoryear{Ulyanov, Vedaldi, and Lempitsky}{Ulyanov
  et~al\mbox{.}}{2018}]%
        {Ulyanov18GEN}
\bibfield{author}{\bibinfo{person}{Dmitry Ulyanov}, \bibinfo{person}{Andrea
  Vedaldi}, {and} \bibinfo{person}{Victor~S. Lempitsky}.}
  \bibinfo{year}{2018}\natexlab{}.
\newblock \showarticletitle{It Takes (Only) Two: Adversarial Generator-Encoder
  Networks}. In \bibinfo{booktitle}{\emph{Proceedings of the Thirty-Second
  {AAAI} Conference on Artificial Intelligence, (AAAI-18)}}.
  \bibinfo{pages}{1250--1257}.
\newblock


\bibitem[\protect\citeauthoryear{Upchurch, Gardner, Pleiss, Pless, Snavely,
  Bala, and Weinberger}{Upchurch et~al\mbox{.}}{2017}]%
        {Upchurch17DFI}
\bibfield{author}{\bibinfo{person}{Paul Upchurch}, \bibinfo{person}{Jacob~R.
  Gardner}, \bibinfo{person}{Geoff Pleiss}, \bibinfo{person}{Robert Pless},
  \bibinfo{person}{Noah Snavely}, \bibinfo{person}{Kavita Bala}, {and}
  \bibinfo{person}{Kilian~Q. Weinberger}.} \bibinfo{year}{2017}\natexlab{}.
\newblock \showarticletitle{Deep Feature Interpolation for Image Content
  Changes}. In \bibinfo{booktitle}{\emph{Proceedings of International
  Conference on Computer Vision and Pattern Recognition}}.
  \bibinfo{pages}{6090--6099}.
\newblock


\bibitem[\protect\citeauthoryear{Wong}{Wong}{1992}]%
        {Wong1992isomorphism}
\bibfield{author}{\bibinfo{person}{E.~K. Wong}.}
  \bibinfo{year}{1992}\natexlab{}.
\newblock \showarticletitle{Model matching in robot vision by subgraph
  isomorphism}.
\newblock \bibinfo{journal}{\emph{Pattern Recognit.}} \bibinfo{volume}{25},
  \bibinfo{number}{3} (\bibinfo{year}{1992}), \bibinfo{pages}{287--303}.
\newblock


\bibitem[\protect\citeauthoryear{Wood, Baltrusaitis, Morency, Robinson, and
  Bulling}{Wood et~al\mbox{.}}{2016}]%
        {wood2016UnityEyes}
\bibfield{author}{\bibinfo{person}{Erroll Wood}, \bibinfo{person}{Tadas
  Baltrusaitis}, \bibinfo{person}{Louis{-}Philippe Morency},
  \bibinfo{person}{Peter Robinson}, {and} \bibinfo{person}{Andreas Bulling}.}
  \bibinfo{year}{2016}\natexlab{}.
\newblock \showarticletitle{Learning an appearance-based gaze estimator from
  one million synthesised images}. In \bibinfo{booktitle}{\emph{{ACM} Symposium
  on Eye Tracking Research {\&} Applications, {ETRA}}}.
  \bibinfo{pages}{131--138}.
\newblock


\bibitem[\protect\citeauthoryear{Wood, Baltrusaitis, Morency, Robinson, and
  Bulling}{Wood et~al\mbox{.}}{2018a}]%
        {wood2018gaze}
\bibfield{author}{\bibinfo{person}{Erroll Wood}, \bibinfo{person}{Tadas
  Baltrusaitis}, \bibinfo{person}{Louis{-}Philippe Morency},
  \bibinfo{person}{Peter Robinson}, {and} \bibinfo{person}{Andreas Bulling}.}
  \bibinfo{year}{2018}\natexlab{a}.
\newblock \showarticletitle{GazeDirector: Fully Articulated Eye Gaze
  Redirection in Video}.
\newblock \bibinfo{journal}{\emph{Comput. Graph. Forum}} \bibinfo{volume}{37},
  \bibinfo{number}{2} (\bibinfo{year}{2018}), \bibinfo{pages}{217--225}.
\newblock


\bibitem[\protect\citeauthoryear{Wood, Baltru{\v{s}}aitis, Morency, Robinson,
  and Bulling}{Wood et~al\mbox{.}}{2018b}]%
        {wood2018gazedirector}
\bibfield{author}{\bibinfo{person}{Erroll Wood}, \bibinfo{person}{Tadas
  Baltru{\v{s}}aitis}, \bibinfo{person}{Louis-Philippe Morency},
  \bibinfo{person}{Peter Robinson}, {and} \bibinfo{person}{Andreas Bulling}.}
  \bibinfo{year}{2018}\natexlab{b}.
\newblock \showarticletitle{GazeDirector: Fully articulated eye gaze
  redirection in video}. In \bibinfo{booktitle}{\emph{Computer Graphics
  Forum}}, Vol.~\bibinfo{volume}{37}. Wiley Online Library,
  \bibinfo{pages}{217--225}.
\newblock


\bibitem[\protect\citeauthoryear{Wood, Baltrusaitis, Zhang, Sugano, Robinson,
  and Bulling}{Wood et~al\mbox{.}}{2015}]%
        {wood2015SynthesEyes}
\bibfield{author}{\bibinfo{person}{Erroll Wood}, \bibinfo{person}{Tadas
  Baltrusaitis}, \bibinfo{person}{Xucong Zhang}, \bibinfo{person}{Yusuke
  Sugano}, \bibinfo{person}{Peter Robinson}, {and} \bibinfo{person}{Andreas
  Bulling}.} \bibinfo{year}{2015}\natexlab{}.
\newblock \showarticletitle{Rendering of Eyes for Eye-Shape Registration and
  Gaze Estimation}. In \bibinfo{booktitle}{\emph{International Conference on
  Computer Vision}}. \bibinfo{pages}{3756--3764}.
\newblock


\bibitem[\protect\citeauthoryear{Xia, Yang, and Xue}{Xia et~al\mbox{.}}{2019}]%
        {xia2019explicit}
\bibfield{author}{\bibinfo{person}{Weihao Xia}, \bibinfo{person}{Yujiu Yang},
  {and} \bibinfo{person}{Jing{-}Hao Xue}.} \bibinfo{year}{2019}\natexlab{}.
\newblock \showarticletitle{Unsupervised Multi-Domain Multimodal Image-to-Image
  Translation with Explicit Domain-Constrained Disentanglement}.
\newblock  (\bibinfo{year}{2019}).
\newblock
\urldef\tempurl%
\url{http://arxiv.org/abs/1911.00622}
\showURL{%
\tempurl}


\bibitem[\protect\citeauthoryear{Yang and Zhang}{Yang and Zhang}{2002}]%
        {yang2002eye}
\bibfield{author}{\bibinfo{person}{Ruigang Yang} {and}
  \bibinfo{person}{Zhengyou Zhang}.} \bibinfo{year}{2002}\natexlab{}.
\newblock \showarticletitle{Eye gaze correction with stereovision for
  video-teleconferencing}. In \bibinfo{booktitle}{\emph{European Conference on
  Computer Vision}}. Springer, \bibinfo{pages}{479--494}.
\newblock


\bibitem[\protect\citeauthoryear{Yu, Liu, and Odobez}{Yu
  et~al\mbox{.}}{2019a}]%
        {odobez2019gaze}
\bibfield{author}{\bibinfo{person}{Yu Yu}, \bibinfo{person}{Gang Liu}, {and}
  \bibinfo{person}{Jean{-}Marc Odobez}.} \bibinfo{year}{2019}\natexlab{a}.
\newblock \showarticletitle{Improving Few-Shot User-Specific Gaze Adaptation
  via Gaze Redirection Synthesis}. In \bibinfo{booktitle}{\emph{{IEEE}
  Conference on Computer Vision and Pattern Recognition}}.
  \bibinfo{pages}{11937--11946}.
\newblock


\bibitem[\protect\citeauthoryear{Yu, Liu, and Odobez}{Yu
  et~al\mbox{.}}{2019b}]%
        {yu2019improving}
\bibfield{author}{\bibinfo{person}{Yu Yu}, \bibinfo{person}{Gang Liu}, {and}
  \bibinfo{person}{Jean-Marc Odobez}.} \bibinfo{year}{2019}\natexlab{b}.
\newblock \showarticletitle{Improving Few-Shot User-Specific Gaze Adaptation
  via Gaze Redirection Synthesis}. In \bibinfo{booktitle}{\emph{Proceedings of
  the IEEE Conference on Computer Vision and Pattern Recognition}}.
  \bibinfo{pages}{11937--11946}.
\newblock


\bibitem[\protect\citeauthoryear{Zhang, Isola, Efros, Shechtman, and
  Wang}{Zhang et~al\mbox{.}}{2018}]%
        {zhang2018perceptual}
\bibfield{author}{\bibinfo{person}{Richard Zhang}, \bibinfo{person}{Phillip
  Isola}, \bibinfo{person}{Alexei~A Efros}, \bibinfo{person}{Eli Shechtman},
  {and} \bibinfo{person}{Oliver Wang}.} \bibinfo{year}{2018}\natexlab{}.
\newblock \showarticletitle{The Unreasonable Effectiveness of Deep Features as
  a Perceptual Metric}. In \bibinfo{booktitle}{\emph{CVPR}}.
\newblock


\bibitem[\protect\citeauthoryear{Zhang, Sugano, Fritz, and Bulling}{Zhang
  et~al\mbox{.}}{2015}]%
        {zhang15_cvpr}
\bibfield{author}{\bibinfo{person}{Xucong Zhang}, \bibinfo{person}{Yusuke
  Sugano}, \bibinfo{person}{Mario Fritz}, {and} \bibinfo{person}{Andreas
  Bulling}.} \bibinfo{year}{2015}\natexlab{}.
\newblock \showarticletitle{Appearance-based Gaze Estimation in the Wild}. In
  \bibinfo{booktitle}{\emph{CVPR}}. \bibinfo{pages}{4511--4520}.
\newblock


\bibitem[\protect\citeauthoryear{Zhang, Sugano, Fritz, and Bulling}{Zhang
  et~al\mbox{.}}{2017}]%
        {Zhang2017gaze}
\bibfield{author}{\bibinfo{person}{Xucong Zhang}, \bibinfo{person}{Yusuke
  Sugano}, \bibinfo{person}{Mario Fritz}, {and} \bibinfo{person}{Andreas
  Bulling}.} \bibinfo{year}{2017}\natexlab{}.
\newblock \showarticletitle{It's Written All Over Your Face: Full-Face
  Appearance-Based Gaze Estimation}. In \bibinfo{booktitle}{\emph{{IEEE}
  Conference on Computer Vision and Pattern Recognition Workshops}}.
  \bibinfo{pages}{2299--2308}.
\newblock


\bibitem[\protect\citeauthoryear{Zhu, Yang, and Xiang}{Zhu
  et~al\mbox{.}}{2011}]%
        {zhu2011eye}
\bibfield{author}{\bibinfo{person}{Jiejie Zhu}, \bibinfo{person}{Ruigang Yang},
  {and} \bibinfo{person}{Xueqing Xiang}.} \bibinfo{year}{2011}\natexlab{}.
\newblock \showarticletitle{Eye contact in video conference via fusion of
  time-of-flight depth sensor and stereo}.
\newblock \bibinfo{journal}{\emph{3D Research}} \bibinfo{volume}{2},
  \bibinfo{number}{3} (\bibinfo{year}{2011}), \bibinfo{pages}{5}.
\newblock


\bibitem[\protect\citeauthoryear{Zhu, Park, Isola, and Efros}{Zhu
  et~al\mbox{.}}{2017a}]%
        {zhu2017unpaired}
\bibfield{author}{\bibinfo{person}{Jun-Yan Zhu}, \bibinfo{person}{Taesung
  Park}, \bibinfo{person}{Phillip Isola}, {and} \bibinfo{person}{Alexei~A
  Efros}.} \bibinfo{year}{2017}\natexlab{a}.
\newblock \showarticletitle{Unpaired image-to-image translation using
  cycle-consistent adversarial networks}. In
  \bibinfo{booktitle}{\emph{Proceedings of the IEEE International Conference on
  Computer Vision}}. \bibinfo{pages}{2223--2232}.
\newblock


\bibitem[\protect\citeauthoryear{Zhu, Zhang, Pathak, Darrell, Efros, Wang, and
  Shechtman}{Zhu et~al\mbox{.}}{2017b}]%
        {zhu2017toward}
\bibfield{author}{\bibinfo{person}{Jun-Yan Zhu}, \bibinfo{person}{Richard
  Zhang}, \bibinfo{person}{Deepak Pathak}, \bibinfo{person}{Trevor Darrell},
  \bibinfo{person}{Alexei~A Efros}, \bibinfo{person}{Oliver Wang}, {and}
  \bibinfo{person}{Eli Shechtman}.} \bibinfo{year}{2017}\natexlab{b}.
\newblock \showarticletitle{Toward multimodal image-to-image translation}. In
  \bibinfo{booktitle}{\emph{Advances in Neural Information Processing
  Systems}}.
\newblock


\end{thebibliography}

\end{document}